\PassOptionsToPackage{table}{xcolor}
\documentclass[11pt]{article}

\usepackage[final]{acl}

\usepackage{times}
\usepackage{latexsym}

\usepackage[T1]{fontenc}

\usepackage[utf8]{inputenc}

\usepackage{microtype}

\usepackage{inconsolata}

\usepackage{graphicx}

\usepackage{booktabs}    
\usepackage{makecell}    
\usepackage{array}       
\usepackage{multirow}    
\usepackage{arydshln}    
\usepackage{caption}     
\usepackage[T1]{fontenc} 
\usepackage{amsmath}     
\usepackage{xcolor}  
\usepackage{colortbl} 
\usepackage{tabularx}
\usepackage{comment}
\usepackage{pgfplots}
\pgfplotsset{compat=1.18}
\usepackage{tikz}
\usepackage{float}
\usepackage{pifont}
\usepackage{makecell}
\usepackage{adjustbox}
\usepackage{amsmath,amssymb}
\usepackage{afterpage}
\usepackage{placeins} 
\usepackage{dblfloatfix} 
\usepackage{listings}
\usepackage{tikz}
\usepackage{pgfplots}
\usepackage{pgfplotstable}
\usepgfplotslibrary{colormaps}
\pgfplotsset{compat=1.18}
\usepgfplotslibrary{groupplots}
\usetikzlibrary{calc} 
\pgfkeys{/tikz/.cd,
  cube top color/.store in=\CubeTopColor,
  cube top color=blue!60,
  cube front color/.store in=\CubeFrontColor,
  cube front color=blue!30,
  cube side color/.store in=\CubeSideColor,
  cube side color=blue!40,
  3d cube color/.code={\colorlet{mycolor}{#1}%
    \tikzset{
      cube top color=mycolor!60,
      cube front color=mycolor!30,
      cube side color=mycolor!40,
      draw=mycolor
    }}
}
\lstdefinestyle{prompt}{
  basicstyle=\ttfamily\footnotesize,
  breaklines=true,
  breakatwhitespace=false,
  columns=fullflexible,
  keepspaces=true,
  frame=single,
  framerule=0.3pt,
  xleftmargin=0pt,
  framexleftmargin=0pt,
  aboveskip=6pt, belowskip=6pt,
  literate={|}{{\textbar{}\allowbreak}}1
}

\newcommand{\rowdash}{\hdashline[0.5pt/2pt]\addlinespace[2pt]}
\newcommand{\cmark}{\ding{51}} 
\newcommand{\xmark}{\ding{55}} 
\newcolumntype{C}[1]{>{\centering\arraybackslash}p{#1}} 
\newcolumntype{L}[1]{>{\raggedright\arraybackslash}p{#1}}

\definecolor{tabfactblue}{RGB}{66,96,137}   
\definecolor{wikitqgold}{RGB}{238,195,128}  

\title{CORE-T: COherent REtrieval of Tables for Text-to-SQL}

\author{
  Hassan Soliman\textsuperscript{1} \quad
  Vivek Gupta\textsuperscript{2} \quad
  Dan Roth\textsuperscript{3,4} \quad
  Iryna Gurevych\textsuperscript{1}
  \\
  \textsuperscript{1}Ubiquitous Knowledge Processing Lab (UKP Lab), Department of Computer Science,\\TU Darmstadt and National Research Center for Applied Cybersecurity ATHENE, Germany\\
  \textsuperscript{2}Arizona State University \quad
  \textsuperscript{3}University of Pennsylvania \quad
  \textsuperscript{4}Oracle AI \\
  \href{https://www.ukp.tu-darmstadt.de}{www.ukp.tu-darmstadt.de}
}

\begin{document}
\maketitle
\begin{abstract}
Realistic text-to-SQL workflows often require joining multiple tables. As a result, accurately retrieving the relevant set of tables becomes a key bottleneck for end-to-end performance. 
We study an \emph{open-book} setting where queries must be answered over large, heterogeneous table collections pooled from many sources, without clean scoping signals such as database identifiers. Here, dense retrieval (DR) achieves high recall but returns many distractors, while join-aware alternatives often rely on extra assumptions and/or incur high inference overhead.
We propose \textsc{CORE-T}, a scalable, training-free framework that enriches tables with LLM-generated purpose metadata and pre-computes a lightweight table-compatibility cache. At inference time, DR returns top-$K$ candidates; a single LLM call selects a coherent, joinable subset, and a two-step additive adjustment stage restores strongly compatible tables.
Across \textsc{Bird}, \textsc{Spider}, \textsc{MMQA}, and \textsc{Beaver}, \textsc{CORE-T} improves over DR by up to 22.7 points in table-selection F1 while returning up to 40\% fewer tables, and by up to 24.4 points in multi-table execution accuracy, and uses 1.64--4.20$\times$ fewer total selection tokens than LLM-intensive baselines.\footnote{Our code and data are available at \url{https://github.com/UKPLab/emnlp2026-core-t}.}
\end{abstract}

\section{Introduction}
\label{sec:introduction}

Natural language interfaces to structured data (text-to-SQL) aim to let non-experts
query relational tables in everyday language \citep{yu-etal-2018-spider, li-etal-2023-bird}.
Modern pipelines often retrieve relevant tables before generating SQL, making table
retrieval a critical bottleneck: if the retrieved set misses required tables, contains
distractors, or lacks a valid join path, the SQL generator must guess joins, drop
constraints, or produce incomplete SQL.

In open-book text-to-SQL, retrieval is not merely a query--table relevance problem.
Many analytical questions require composing evidence across normalized tables, so the
retriever must return a \emph{set} that is both semantically relevant and structurally
coherent. Relevance-only retrieval can hand the SQL generator plausible but
disconnected tables rather than a compact, join-ready schema slice.

\begin{figure}[t]
  \centering
  \includegraphics[width=\columnwidth]{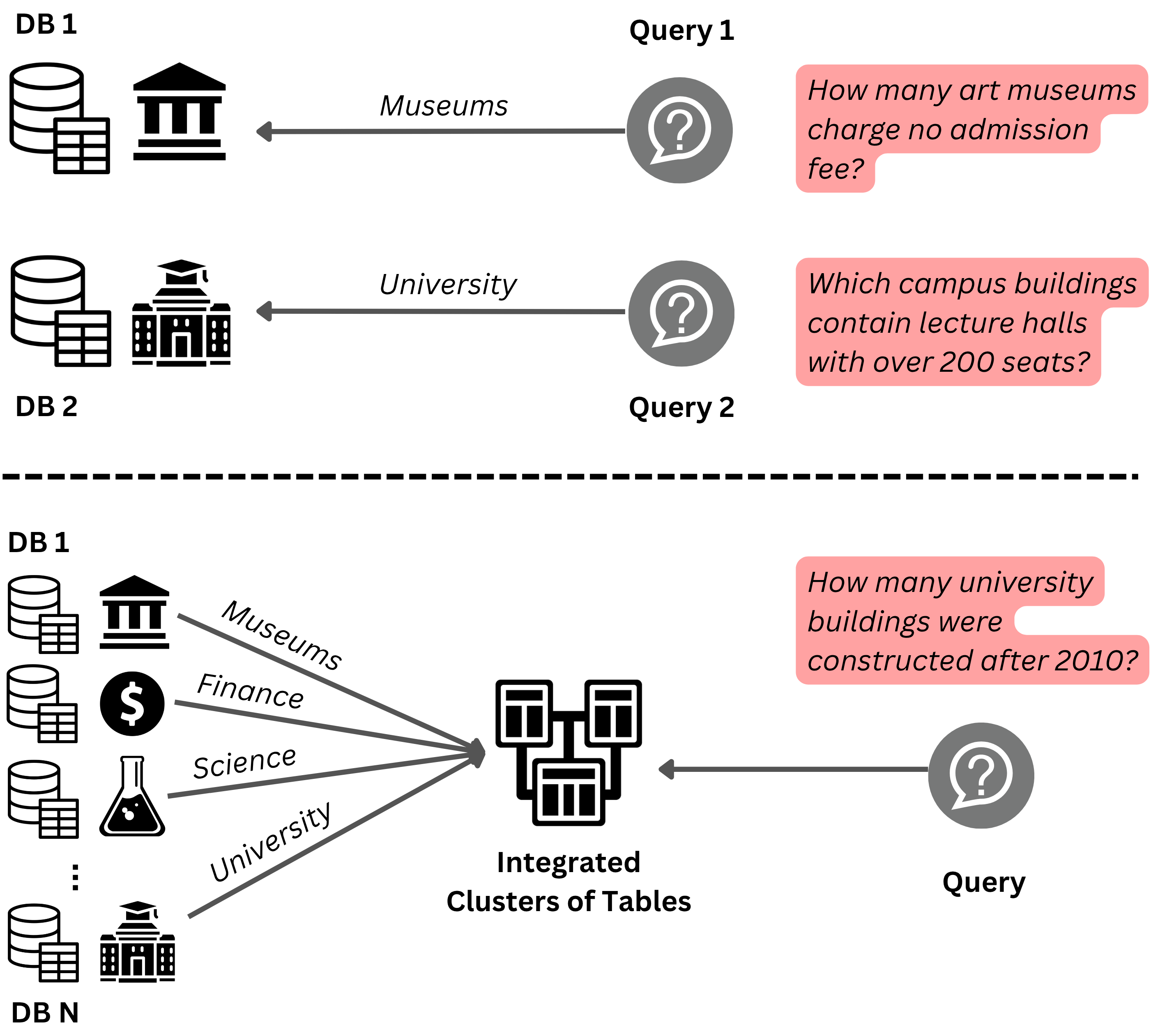}
  \caption{\textbf{Querying setups.}
  \emph{Top:} Closed-book: each query targets a known database.
  \emph{Bottom:} Open-book: queries must be answered over \emph{integrated} clusters of tables spanning various domains pooled from multiple database schemas.}
  \label{fig:setup-contrast}
\end{figure}

Most prior text-to-SQL and table-retrieval work assumes a \emph{closed-book} setting,
where the target database or schema graph is known and retrieval is confined to a
small schema \citep{yu-etal-2018-spider,lee-etal-2021-kaggledbqa,li-etal-2023-bird,herzig-etal-2021-open,chen2020tablesearch,wang-etal-2022-table}.
In contrast, data-lake and semantic join discovery research studies \emph{open-book}
analytics over large heterogeneous table collections, where relevant and joinable
tables must be discovered without a predefined schema graph
\citep{zhu2019josie,nargesian2019datalake,zhang2020findingrelated,cong-etal-2023-warpgate}.
This mirrors integrated enterprise corpora, where tables are pooled from many sources
and clean scoping signals such as database identifiers (\texttt{db\_id}) are unavailable.

Figure~\ref{fig:setup-contrast} illustrates the challenge. In a closed-book setting (top),
queries such as \emph{``How many art museums charge no admission fee?''} can be
answered within a delimited schema (\texttt{db\_id}), e.g., museums vs.\ university. In an open-book
setting (bottom), queries such as \emph{``How many university buildings were
constructed after 2010?''} require retrieval over pooled tables from many domains.
A table named \texttt{buildings} may occur in both museum and university collections;
without \texttt{db\_id}, the retriever must disambiguate candidates using attributes
and relationships. Missing one bridge table can break the join path, while
similar-looking distractors can induce spurious joins; high recall alone is therefore
insufficient.

Our goal is therefore not to introduce a new retrieval model in isolation, but to provide a \emph{training-free systems design} for \emph{pooled open-book} corpora where \texttt{db\_id} and gold foreign keys are unavailable, and where the central bottleneck is \emph{join-coherent} table-set retrieval. Existing retrievers address parts of this goal but leave important gaps. Dense retrieval
(DR)~\citep{karpukhin2020dense} scales well but
remains join-agnostic. ReAct~\citep{yao-etal-2023-react} can iteratively expand
evidence but requires multiple LLM calls. Join-aware methods such as JAR and
ARM~\citep{chen-etal-2024-table,chen-etal-2025-retrieve} model relational structure,
but rely on scoping assumptions such as \texttt{db\_id} and/or incur substantial
inference overhead. REAR~\citep{agarwal2025rear} improves multi-table retrieval via
retrieve--expand--refine stages, but column-similarity-based join evidence can produce false positives in pooled corpora with semantically similar yet non-joinable
tables.

\begin{figure}[!tbp]
  \centering
  \includegraphics[width=\columnwidth]{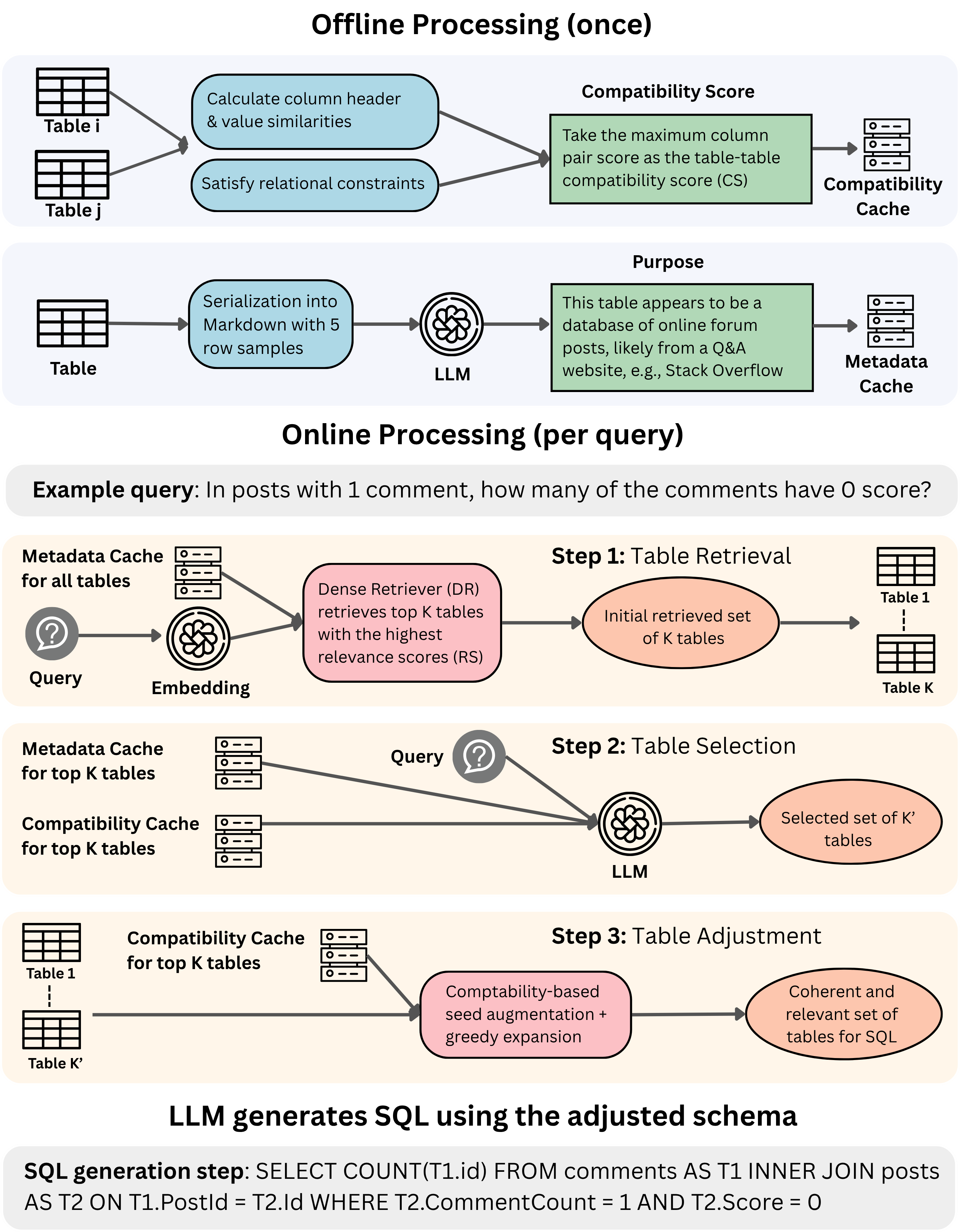}
  \caption{\textbf{Overview.} CORE-T combines offline table enrichment and compatibility caching with a lightweight online pipeline: DR, single-shot LLM selection, and a two-step additive adjustment final stage.}
  \label{fig:main-figure}
\end{figure}

This raises our central question: \emph{Can we design a scalable, training-free
retriever for pooled open-book corpora that jointly exploits query--table relevance
and table--table compatibility, without assuming \texttt{db\_id} or gold foreign keys?}

We propose \textsc{CORE-T}, a framework for \emph{COherent REtrieval of Tables} for
text-to-SQL. As shown in Figure~\ref{fig:main-figure}, \textsc{CORE-T} moves reusable
schema understanding offline and keeps online inference lightweight: it enriches tables
with LLM-generated purpose metadata, precomputes a compatibility cache from
column-level semantic and value-based signals, retrieves a high-recall top-$K$
candidate set, uses one LLM call to select a coherent subset, and applies a two-step
additive adjustment to restore strongly compatible and over-pruned candidates. In summary, we contribute:

\paragraph{Training-free join-coherent table-set retrieval for pooled open-book corpora.}
  We introduce \textsc{CORE-T}, a scalable retrieval layer that couples
  purpose-enriched query--table retrieval with cached table--table compatibility
  evidence to produce compact, join-coherent schema slices. Compatibility provides
  structured guidance for LLM subset selection and targeted additive recovery, enabling
  open-book retrieval without \texttt{db\_id} scoping or gold foreign-key annotations.

\paragraph{Improved effectiveness and efficiency under pooled evaluation.}
  We compare against DR, ReAct, and recent join-aware methods (JAR/ARM/REAR)
  under pooled multi-database evaluation on \textsc{Bird}, \textsc{Spider},
  \textsc{MMQA}, and \textsc{Beaver}~\citep{yu-etal-2018-spider,li-etal-2023-bird,wu-etal-2025-MMQA,chen-etal-2025-beaver}.
  \textsc{CORE-T} improves the precision--recall balance for multi-table retrieval,
  returns more coherent table sets for SQL generation, and reduces LLM usage
  (up to $\sim$5$\times$ fewer input tokens than heavier multi-draft LLM generation methods, e.g., ARM).

\section{Related Work and Positioning}
Given the pooled open-book setting introduced in Section~\ref{sec:introduction}, the key distinction among prior methods is not only whether they retrieve relevant tables, but also how they maintain table-set coherence when join evidence is noisy and scoping information is limited. This distinction is especially important in integrated corpora, where noisy join signals and near-duplicate cross-domain tables make selection difficult. JAR and ARM build join graphs and use MIP-based optimization, making them sensitive to the induced compatibility structure; both assume \texttt{db\_id} for scoping, and ARM adds LLM self-verification overhead. REAR is LLM-free at online inference, separating semantic retrieval from joinability-based expansion and refinement, but its local, column-similarity-driven join evidence can be brittle with semantically similar yet non-joinable columns. In contrast, \textsc{CORE-T} targets open-book corpora with (i) LLM-generated \emph{purpose} metadata for candidate disambiguation without \texttt{db\_id}, (ii) modified compatibility scoring for more reliable join evidence, and (iii) single-shot LLM selection and a two-step additive adjustment stage to retain a high-recall, join-coherent table set.

\section{Overview}

Figure~\ref{fig:main-figure} provides an overview: schema understanding is moved to offline enrichment and caching, while online inference remains lightweight.

\paragraph{Offline.} We enrich tables with brief purpose metadata and build a dense index over enriched table representations, following offline index enrichment \citep{chen2025enrichindex}. We also pre-compute a table--table compatibility cache that approximates joinability and provides candidate join edges.

\paragraph{Online.} Given a query, DR returns a top-$K$ candidate set. A single LLM call selects a coherent, connected subset using the candidate tables and cached compatibility evidence, and a two-step additive adjustment recovers strongly compatible tables from the original top-$K$ set before SQL generation.

\section{Methodology}
\label{sec:methodology}

We propose a scalable, join-aware multi-table retriever for \emph{open-book} text-to-SQL, where tables from multiple DBs are pooled, and the system must retrieve relevant, joinable tables \emph{without} \texttt{db\_id}s or gold foreign keys. Let $\mathcal{T}=\{t_1,\ldots,t_N\}$ be the unified table corpus. Given a query $q$, we output a table set $S(q)\subseteq \mathcal{T}$ that balances high gold-table recall with fewer irrelevant tables, while remaining coherent for downstream SQL generation.

\subsection{Offline Processing and Caching}
\label{sec:offline-processing}

Our method uses two offline, reusable, query-agnostic signals: (i) an enriched table index for DR, used to compute the relevance score $\mathrm{RS}(q,t)$ at inference time, and (ii) a table--table compatibility score $\mathrm{CS}(t_i,t_j)$ that approximates joinability.

\paragraph{Table enrichment and indexing.}
Offline, each table $t$ is serialized as a 5-row Markdown representation and augmented with an LLM-generated \emph{purpose description}. We sample 5 rows uniformly without replacement to provide a lightweight value snapshot, avoiding full-table inputs while exposing different values and formats. The purpose description summarizes the table's contents and typical use (e.g., key entities, attributes, and granularity), using a fixed prompt in Appendix~\ref{app:prompts} (cf.\ \S\ref{lst:prompt-purpose}). We embed the concatenation of the Markdown snapshot and purpose as $e_t=f_{\text{tbl}}(\texttt{Markdown+purpose})$ and store all vectors in a FAISS index for online retrieval. Appendix~\ref{app:table-enrichment-details} discusses our compact metadata design and 5-row cap.

\paragraph{Column signals.}
To approximate joinability without foreign keys, we compute lightweight column-level signals. For each column $c$, we embed its header text (\texttt{table\_name}+\texttt{column\_name}) with $f_{\text{col}}$ to obtain $e_c$. For each cross-table pair $(c_i,c_j)$, we compute (i) header similarity (exact lexical + embedding-based semantic), (ii) value overlap (Jaccard), inspired by the pairwise similarity signals used in JAR, and two \emph{relational constraints} that better mimic key--foreign-key joins: (iii) \emph{uniqueness} and (iv) \emph{subset}, all ignoring \textsc{null}s.

\paragraph{Compatibility cache.}
We combine these signals into a column-pair compatibility score $s(c_i,c_j)\in[0,1]$ with a hand-crafted function and a hard key--foreign-key-like constraint: we only score pairs where \emph{at least one} column is unique and the values exhibit a subset relation. This promotes key-like joins while suppressing spurious matches from generic columns (e.g., \texttt{id}, \texttt{name}) in pooled corpora. Details and the scoring function are in Appendix~\ref{app:compat-eval} (Eq.~\ref{eq:colscore}, Figure~\ref{fig:compatibility-scores}). The table--table compatibility score is the best valid column match:
{%
\setlength{\abovedisplayskip}{2pt}
\setlength{\belowdisplayskip}{2pt}
\setlength{\abovedisplayshortskip}{2pt}
\setlength{\belowdisplayshortskip}{2pt}
\begin{equation}
\mathrm{CS}(t_i,t_j)=
\max_{\substack{c\in C(t_i),\,c'\in C(t_j)\\ \text{valid}(c,c')}} s(c,c'),
\end{equation}
}%
and we cache the corresponding argmax column pair; if no valid pair exists, $\mathrm{CS}(t_i,t_j)=0$. Each compatibility edge encodes a single-column equi-join, but Selection and Adjustment can compose multiple edges into multi-table join paths. The cache is computed offline and reused across queries. When reliable values are unavailable, Retrieval and Selection stages remain operational, while compatibility-guided Adjustment steps avoids unsupported expansion. Appendix~\ref{app:compat-eval} reports gains over JAR-style similarity scoring (Table~\ref{tab:compatibility-eval}) and discusses sparse inference-time construction and incremental maintenance as the corpus changes.

\subsection{Online Processing and Inference}
\label{sec:online-processing}

Given a query $q$, we produce a final table set $S(q)$ in three stages:
(1) DR over enriched table embeddings to obtain top-$K$ candidates,
(2) a single LLM call to select a coherent, joinable subset, and
(3) a two-step additive adjustment stage that restores strong compatible tables omitted by the selector.

\subsubsection{Table Retrieval}
\label{sec:table-retrieval}

We obtain a high-recall candidate set using DR over enriched table embeddings created offline. This step is illustrated in Appendix~\ref{app:table-selection} (Figure~\ref{fig:initial-table-retrieval}).

\paragraph{Query embedding and scores.}
For each query $q$, we encode it with the table embedding model, $e_q = f_{\text{tbl}}(q)$, and compute its relevance to each table $t$ as cosine similarity, $\mathrm{RS}(q,t)=\cos(e_q,e_t)$.

\paragraph{Top-$K$ initial set.}
We retrieve the top-$K$ tables by $\mathrm{RS}(q,t)$, $T_K(q) = \{t^{(1)}, \dots, t^{(K)}\}$.
This set is high-recall but may include loosely related or distractor tables, motivating selection and adjustment stages. Subsequent stages operate on $T_K(q)$.

\subsubsection{Table Selection}
\label{sec:table-selection}

DR optimizes query--table relevance but ignores table--table interactions. To obtain a smaller, joinable subset while preserving high recall, we use a single LLM call prompted as a \emph{SQL schema analyst} that follows a human-reasoning workflow to jointly reason over the query, candidate tables, and cached compatibility evidence. The prompt is \emph{few-shot}, with one synthetic example illustrating the expected input and output format. Full details are in Appendix~\ref{app:table-selection} (cf.\ \S\ref{lst:prompt-selection}).

\paragraph{LLM input and output.}
Given $T_K(q)=\{t^{(1)},\dots,t^{(K)}\}$, we provide: (i) the query $q$; (ii) an indexed list of the $K$ candidate tables (name, 5-row Markdown snapshot, generated purpose); and (iii) compatibility evidence for pairs with $\mathrm{CS}(t_i,t_j)>0$, including \texttt{overall\_compatibility} score and \texttt{best\_join\_columns}. Pairs not listed are treated as having no join edge. The LLM forms connected groups and selects one, parsed from JSON as $T_{K'}(q)\subseteq T_K(q)$. This subset is typically smaller and more coherent due to conservative pruning instructions, but may over-prune compatible tables. We address this with the compatibility-driven adjustment step that restores strongly compatible ignored tables from the original top-$K$ set.

\subsubsection{Table Adjustment}
\label{sec:table-adjustment}

\begin{figure}[!tbp]
  \centering
  \includegraphics[width=\columnwidth]{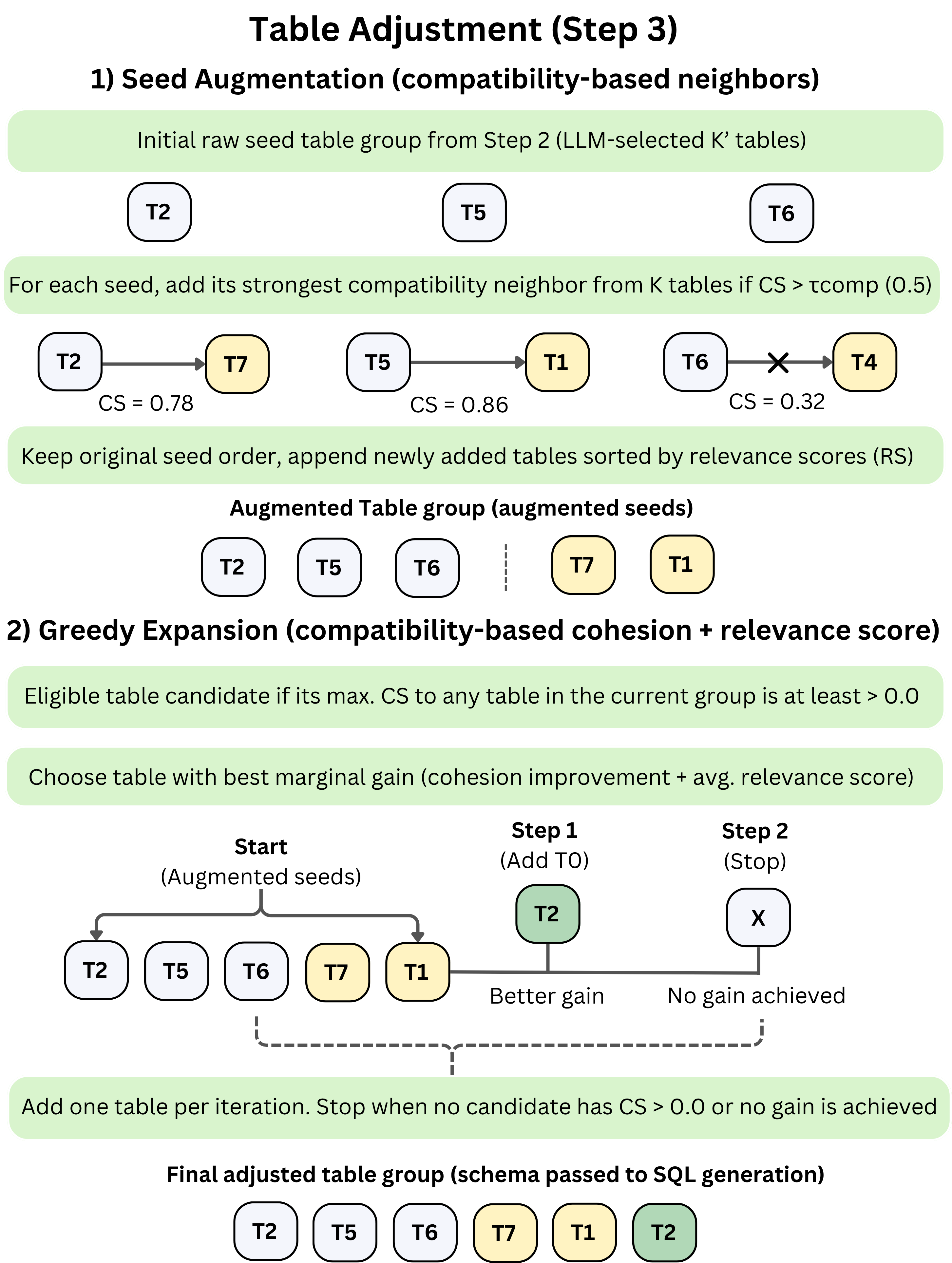}
  \caption{\textbf{Table Adjustment.} The LLM-selected seed tables are first augmented with their strongest compatible neighbors, preserving seed order and ranking added tables by retriever relevance. The group is then expanded greedily by adding eligible candidates with the largest marginal gain until no beneficial table remains.}
  \label{fig:table-adjustment}
\end{figure}

\paragraph{Seeded compatibility-based augmentation.}
As illustrated in Figure~\ref{fig:table-adjustment}, the adjustment step treats
the LLM-selected tables \(T_{K'}(q)\) as a seed group within the retrieved
top-\(K\) candidates \(T_K(q)\). For each seed table, we add its strongest
compatibility neighbor in \(T_K(q)\), according to cached \(\mathrm{CS}\), only
when the score exceeds \(\tau_{\mathrm{comp}}\). The resulting augmented group \(G_0\) preserves the LLM-selected order and appends newly added tables sorted by descending relevance score to the query. \(G_0\) then initializes the greedy expansion.

\paragraph{Greedy expansion.}
Starting with \(G\leftarrow G_0\), we greedily add tables from
\(T_K(q)\setminus G\) until no candidate yields positive marginal gain.
A candidate is eligible only if it has nonzero compatibility with the current
group. Among eligible candidates, we select the table with the largest marginal
gain, defined as the combined improvement in mean pairwise compatibility plus
mean relevance score to the query. The final schema \(S(q)\) preserves
the LLM-selected seeds while recovering additional compatible and relevant
tables for SQL generation. Full details and equations are in
Appendix~\ref{app:table-adjustment}.

\section{Experiments}

\subsection{Experimental Setup}

\subsubsection{Datasets}
\label{sec:datasets}

\begin{table}[t]
    \centering
    \scriptsize
    \setlength{\tabcolsep}{2pt}
    \renewcommand{\arraystretch}{0.95}
    \resizebox{\columnwidth}{!}{%
    \begin{tabular}{@{}lrrrrrrrrr@{}}
    \toprule
    Dataset & \#DB & \#Tab & \#Empty & Rows/T & Cols/T & \#Q & MT\% & EG\% & $\bar{G}$ \\
    \midrule
    BIRD   & 11 &  75 &   0 & 52{,}437 & 10.6 & 1{,}534 & 76.4 &  0.0 & 1.95 \\
    SPIDER & 20 &  81 &   0 &  6{,}665 &  5.4 & 1{,}034 & 44.4 &  4.7 & 1.51 \\
    MMQA         &  1 & 710 &   0 &  1{,}547 &  5.1 & 1{,}105 & 99.6 & 10.9 & 2.20 \\
    BEAVER       &  6 & 463 & 214 & 42{,}404 &  9.2 &   209 & 98.6 & 20.1 & 4.44 \\
    \bottomrule
    \end{tabular}%
    }
    \caption{Dataset stats after pooling. \#Empty: zero-row tables; Rows/T and Cols/T: averages per table; MT\%: queries with $\ge2$ gold tables; EG\%: queries with empty gold execution result; $\bar{G}$: avg. gold tables/query.}
    \label{tab:dataset-stats}
\end{table}

We evaluate on the dev splits of \textsc{Bird}~\citep{li-etal-2023-bird} and \textsc{Spider}~\citep{yu-etal-2018-spider}, as well as \textsc{MMQA}~\citep{wu-etal-2025-MMQA} and \textsc{BEAVER}~\citep{chen-etal-2025-beaver}, all with multi-table requirements. \textsc{Bird}/\textsc{Spider} enable comparison with join-aware retrievers (e.g., JAR/ARM), while \textsc{MMQA} and \textsc{BEAVER} provide larger, multi-table-intensive stress tests under integrated corpora. For \textsc{Bird}, \textsc{Spider}, and \textsc{MMQA}, we pool tables from multiple DBs or question-specific schemas into one retrieval corpus and remove \texttt{db\_ids}. \textsc{BEAVER} is already open-book and only requires removing \texttt{db\_id}. Table~\ref{tab:dataset-stats} reports the resulting statistics. For \textsc{MMQA}, we use a stratified one-third subset (1{,}105 queries) for efficiency. Appendix~\ref{app:dataset-preprocessing} provides preprocessing details, benchmark-realism discussion, and notes on additional benchmarks such as \textsc{Spider 2.0}~\citep{lei-etal-2025-spider2}.

\subsubsection{Baselines}
\label{sec:baselines}

We compare against dense retrieval (DR)~\citep{karpukhin2020dense}, an agentic ReAct-based retriever~\citep{yao-etal-2023-react}, and recent join-aware methods: JAR~\citep{chen-etal-2024-table}, ARM~\citep{chen-etal-2025-retrieve}, and REAR~\citep{agarwal2025rear}. All are evaluated in the same open-book setting.

\paragraph{Baseline overview.}
DR retrieves top-$K$ tables by cosine similarity between the query embedding and enriched table embeddings (Markdown + purpose), denoted DR@$K$. ReAct iteratively queries the same dense index for up to three steps. JAR is a join-aware reranker that uses a mixed-integer program (MIP) to select a connected set of $K$ tables by jointly optimizing query coverage and inferred join compatibility. ARM adds LLM-guided alignment for candidate retrieval, followed by join-aware MIP selection and LLM self-verification via multiple draft generations. REAR is an LLM-free retrieve--expand--refine pipeline that retrieves query-relevant base tables, expands them with structurally joinable candidates using precomputed column embeddings, and refines the pool by jointly scoring query--table relevance and table--table joinability. Appendix~\ref{app:baselines} gives implementation details and a thorough comparison; Table~\ref{tab:efficiency-bird-simple} summarizes LLM-call cost, \texttt{db\_id} assumptions, and each method's core \emph{rationale}.

\subsubsection{Evaluation Metrics}
\label{sec:evaluation-metrics}

We evaluate (i) table retrieval/selection quality, (ii) end-to-end text-to-SQL
execution accuracy, and (iii) efficiency. We report set-based precision (P), F1, and perfect recall (PR) for table selection; execution accuracy (EX)
for SQL generation overall (EX$_{all}$), on multi-table queries (EX$_{MT}$), and on the perfect-recall subset (EX$_{PR}$). We also report token-based measures for LLM usage.
More details about metric definitions are provided in Appendix~\ref{app:eval-metrics}.

\subsubsection{Implementation Details}
\label{sec:implementation-details}

\paragraph{LLM setup.}
We use two LLMs for table selection: Llama\mbox{-}3.1\mbox{-}8B\mbox{-}Instruct~\citep{meta2024llama31,grattafiori2024llama3} and Qwen\mbox{-}2.5\mbox{-}7B\mbox{-}Instruct~\citep{qwen2024qwen25}.

\paragraph{Embeddings and thresholds.}
After comparing embedding models on MTEB~\citep{muennighoff-etal-2023-mteb} and its leaderboard, we use UAE\mbox{-}Large\mbox{-}V1 for our initial table retrieval step and all baselines involving a dense-retrieval step~\citep{li-li-2024-aoe}.

\paragraph{Baselines code.}
For JAR and ARM, we use the authors' released code and default hyperparameters on supported datasets: JAR provides reproducible scripts for \textsc{Bird} and \textsc{Spider}, while ARM supports \textsc{Bird}. For REAR, we use the authors' predicted tables from their best reported configuration. More details and discussion are in Appendix~\ref{app:implementation-details}.

\subsection{Results and Analysis}

\begin{table*}[t]
    \centering
    \small
    \setlength{\tabcolsep}{3pt}
    \renewcommand{\arraystretch}{1.1}
    \begin{adjustbox}{max width=\textwidth}
    \begin{tabular}{
        @{} L{1.75cm}
        C{1.15cm} C{0.50cm} C{0.52cm} C{0.52cm}
        >{\centering\arraybackslash}m{4pt}
        C{1.15cm} C{0.50cm} C{0.52cm} C{0.52cm}
        >{\centering\arraybackslash}m{4pt}
        C{1.15cm} C{0.50cm} C{0.52cm} C{0.52cm}
        >{\centering\arraybackslash}m{4pt}
        C{1.15cm} C{0.50cm} C{0.52cm} C{0.52cm}
        @{}
    }
    & \multicolumn{4}{c}{\textbf{Bird} (n=1534, $\bar G$=1.95)} && \multicolumn{4}{c}{\textbf{Spider} (n=1034, $\bar G$=1.51)} && \multicolumn{4}{c}{\textbf{MMQA} (n=1105, $\bar G$=2.20)} && \multicolumn{4}{c}{\textbf{Beaver} (n=209, $\bar G$=4.44)} \\
    \cmidrule(lr){2-5} \cmidrule(lr){7-10} \cmidrule(lr){12-15} \cmidrule(lr){17-20}
    \textbf{} & {\scriptsize Avg.\#tab.$\downarrow$} & \textbf{P} & \textbf{F1} & \textbf{PR}
    & & {\scriptsize Avg.\#tab.$\downarrow$} & \textbf{P} & \textbf{F1} & \textbf{PR}
    & & {\scriptsize Avg.\#tab.$\downarrow$} & \textbf{P} & \textbf{F1} & \textbf{PR}
    & & {\scriptsize Avg.\#tab.$\downarrow$} & \textbf{P} & \textbf{F1} & \textbf{PR} \\
    \midrule
    \multicolumn{20}{l}{\textbf{Llama-3.1-8B-Instruct}} \\
    \midrule
    DR@5 & 5.0 & 34.9 & 49.4 & 83.0 && 5.0 & 29.4 & 43.9 & 95.6  && \textbf{5.0} & \underline{30.8} & \underline{42.6} & 50.3  && \textbf{5.0} & \underline{33.9} & \underline{36.5} & \underline{13.9} \\
    ReAct & \underline{4.8} & \underline{43.1} & \underline{55.6} & 86.1 && \underline{4.5} & \underline{39.7} & \underline{52.2} & 92.8  && 6.1 & 29.6 & 40.4 & \underline{53.1}  && 8.4 & 9.3 & 12.2 & 8.1 \\
    JAR@5 & 5.0 & 35.5 & 50.1 & 86.0 && 5.0 & 29.6 & 44.2 & \textbf{96.7}  && — & — & — & —  && — & — & — & — \\
    REAR & 5.0 & 36.1 & 51.0 & 88.3 && 5.0 & 28.2 & 42.1 & 90.8  && \textbf{5.0} & 28.4 & 39.2 & 47.8  && \textbf{5.0} & 33.6 & 35.5 & 8.1 \\
    ARM & 5.3 & 40.4 & 53.5 & \textbf{90.9} && — & — & — & —  && — & — & — & —  && — & — & — & — \\
    \rowdash
    CORE-T & \textbf{4.1} & \textbf{50.0} & \textbf{62.3} & \underline{90.0} && \textbf{4.2} & \textbf{40.2} & \textbf{53.8} & \underline{96.6}  && \underline{5.1} & \textbf{38.4} & \textbf{49.2} & \textbf{61.3}  && \underline{5.1} & \textbf{39.3} & \textbf{39.5} & \textbf{15.3} \\
    \addlinespace[0.5pt]
    \midrule
    \multicolumn{20}{l}{\textbf{Qwen-2.5-7B-Instruct}} \\
    \midrule
    DR@5 & 5.0 & 35.2 & 49.8 & 84.2 && 5.0 & 29.6 & 44.1 & \textbf{96.4}  && 5.0 & 31.1 & 43.0 & 51.0  && \textbf{5.0} & 34.8 & 37.3 & 14.8 \\
    ReAct & \underline{3.3} & \textbf{65.6} & \underline{71.6} & 81.3 && \textbf{2.7} & \textbf{67.6} & \textbf{75.5} & 93.4  && \underline{4.9} & \underline{42.1} & \underline{50.4} & 55.4  && 8.2 & 21.0 & 25.3 & \underline{16.3} \\
    JAR@5 & 5.0 & 35.5 & 50.1 & \underline{86.0} && 5.0 & 29.5 & 44.1 & \underline{96.3}  && — & — & — & —  && — & — & — & — \\
    REAR & 5.0 & 36.1 & 51.0 & \textbf{88.3} && 5.0 & 28.2 & 42.1 & 90.8  && 5.0 & 28.4 & 39.2 & 47.8  && \textbf{5.0} & 33.6 & 35.5 & 8.1 \\
    ARM & 3.5 & 59.2 & 68.3 & 84.6 && — & — & — & —  && — & — & — & —  && — & — & — & — \\
    \rowdash
    CORE-T & \textbf{3.1} & \underline{63.1} & \textbf{72.3} & \underline{87.0} && \underline{3.0} & \underline{54.4} & \underline{66.8} & 94.9  && \textbf{4.1} & \textbf{48.3} & \textbf{56.4} & \textbf{59.4}  && \underline{5.3} & \textbf{42.8} & \textbf{43.0} & \textbf{17.7} \\
    \addlinespace[0.5pt]
    
    \bottomrule
    \end{tabular}%
    \end{adjustbox}
    \caption{\textbf{Table-selection performance.}
    Avg.\ \#tables$\downarrow$, precision (P), F1, and perfect recall (PR) on \textsc{Bird}, \textsc{Spider}, \textsc{MMQA}, and \textsc{Beaver} (\(\bar G\): avg.\ gold tables/query). All methods use UAE-Large-V1 for dense retrieval; results are grouped by selector LLM.}
    \label{tab:arm-db-id-compact}
\end{table*}

\paragraph{How does the choice of embedding model and top-$K$ affect initial retrieval?}
Table~\ref{tab:dr-topk-results} (Appendix) compares embedding models and top-$K$ cutoffs for initial dense retrieval, including snowflake\mbox{-}arctic\mbox{-}embed\mbox{-}m\mbox{-}v2.0~\citep{yu-etal-2024-arcticembed2}. Overall, text\mbox{-}embedding\mbox{-}3\mbox{-}large is strongest, while UAE\mbox{-}Large\mbox{-}V1 is a close open-source alternative on \textsc{Bird} and \textsc{Spider}; for example, at $K{=}10$ on \textsc{Spider}, both reach 25.7 F1. The gap is larger on \textsc{MMQA} and \textsc{Beaver}. Since increasing $K$ improves PR but lowers precision/F1 and increases context size, we use UAE\mbox{-}Large\mbox{-}V1 with DR@10 as a high-recall starting point.


\begin{table*}[t]
    \centering
    \small
    \setlength{\tabcolsep}{2pt}
    \renewcommand{\arraystretch}{1.1}
    \begin{adjustbox}{max width=\textwidth}
    \begin{tabular}{
      @{} L{3cm}
      c c c
      >{\centering\arraybackslash}m{6pt}
      c c c
      >{\centering\arraybackslash}m{6pt}
      c c c
      >{\centering\arraybackslash}m{6pt}
      c c c
      @{}
    }
    & \multicolumn{3}{c}{\textbf{Bird} (n=1534)} && \multicolumn{3}{c}{\textbf{Spider} (n=1034)} && \multicolumn{3}{c}{\textbf{MMQA} (n=1105)} && \multicolumn{3}{c}{\textbf{Beaver} (n=209)} \\
    \cmidrule(lr){2-4} \cmidrule(lr){6-8} \cmidrule(lr){10-12} \cmidrule(l){14-16}
     & \makecell[c]{EX$_{MT}$\\(76.4\%)} & \makecell[c]{EX$_{all}$\\(100\%)} & \makecell[c]{EX$_{PR}$\\(90.0\%)} && \makecell[c]{EX$_{MT}$\\(44.4\%)} & \makecell[c]{EX$_{all}$\\(100\%)} & \makecell[c]{EX$_{PR}$\\(96.6\%)} && \makecell[c]{EX$_{MT}$\\(99.6\%)} & \makecell[c]{EX$_{all}$\\(100\%)} & \makecell[c]{EX$_{PR}$\\(61.3\%)} && \makecell[c]{EX$_{MT}$\\(98.6\%)} & \makecell[c]{EX$_{all}$\\(100\%)} & \makecell[c]{EX$_{PR}$\\(15.3\%)} \\
    \midrule
    \multicolumn{16}{l}{\textbf{Llama-3.2-3B}} \\
    \midrule
    DR@5 & 3.3 & 4.0 & 3.9 && 10.0 & 13.9 & 13.9 && 2.1 & 2.1 & 1.8 && \underline{0.5} & \underline{0.5} & \underline{0.0} \\
    ReAct & \underline{9.9} & \underline{10.8} & \underline{10.4} && 12.4 & 20.0 & 20.0 && 2.5 & 2.5 & 2.2 && 0.0 & 0.0 & \underline{0.0} \\
    JAR@5 & 7.8 & 8.3 & 8.1 && 10.9 & 14.4 & 14.4 && — & — & — && — & — & — \\
    REAR & 9.6 & 10.0 & 9.8 && \underline{26.4} & \underline{29.6} & \underline{29.3} && \underline{8.4} & \underline{8.3} & \underline{6.9} && \textbf{1.0} & \textbf{1.0} & \underline{0.0} \\
    ARM & 4.9 & 5.1 & 5.0 && — & — & — && — & — & — && — & — & — \\
    \rowdash
    CORE-T & \textbf{15.7} & \textbf{16.6} & \textbf{16.2} && \textbf{34.4} & \textbf{34.5} & \textbf{34.1} && \textbf{18.9} & \textbf{18.8} & \textbf{15.8} && \underline{0.5} & \underline{0.5} & \textbf{0.5} \\
    \rowdash
    Oracle & 24.8 & 27.8 & 27.8 && 45.1 & 58.9 & 58.9 && 45.4 & 45.4 & 45.4 && 1.5 & 1.9 & 1.9 \\
    \addlinespace[4pt]
    \midrule
    \multicolumn{16}{l}{\textbf{Gemma-3-4B}} \\
    \midrule
    DR@5 & 14.6 & 20.0 & 19.0 && 38.8 & 51.5 & 51.1 && \underline{18.9} & \underline{18.9} & \underline{16.8} && \textbf{5.3} & \textbf{5.3} & \textbf{1.0} \\
    ReAct & 15.3 & 21.2 & 20.5 && 40.1 & \underline{51.8} & \underline{51.5} && 18.0 & 17.9 & 15.2 && 3.4 & 3.3 & \underline{0.5} \\
    JAR@5 & \textbf{18.3} & \textbf{23.0} & \textbf{22.2} && \underline{41.4} & 51.4 & 51.3 && — & — & — && — & — & — \\
    REAR & 15.9 & 20.1 & 19.6 && 35.3 & 45.1 & 44.7 && 16.1 & 16.1 & 12.9 && \underline{4.9} & \underline{4.8} & \underline{0.5} \\
    ARM & \underline{16.6} & \underline{22.3} & \underline{21.7} && — & — & — && — & — & — && — & — & — \\
    \rowdash
    CORE-T & \underline{16.6} & 21.9 & 21.1 && \textbf{43.8} & \textbf{54.2} & \textbf{53.8} && \textbf{22.8} & \textbf{22.8} & \textbf{20.8} && 3.9 & 3.8 & \textbf{1.0} \\
    \rowdash
    Oracle & 24.7 & 30.4 & 30.4 && 53.8 & 65.7 & 65.7 && 47.6 & 47.5 & 47.5 && 4.4 & 4.8 & 4.8 \\
    \addlinespace[4pt]
    \midrule
    \multicolumn{16}{l}{\textbf{GPT-4o-mini}} \\
    \midrule
    DR@5 & 34.6 & 40.0 & 37.7 && 53.2 & 65.4 & 64.7 && 31.6 & 31.7 & 27.8 && \textbf{5.3} & \underline{5.3} & \textbf{1.9} \\
    ReAct & 38.1 & \underline{42.5} & 40.9 && 50.5 & 63.7 & 63.4 && \underline{33.2} & \underline{33.1} & \underline{28.7} && 3.4 & 3.3 & \underline{0.5} \\
    JAR@5 & 36.8 & 41.6 & 40.4 && \underline{55.3} & \underline{66.0} & \underline{65.7} && — & — & — && — & — & — \\
    REAR & \underline{38.4} & 42.3 & 40.6 && 52.3 & 62.3 & 61.7 && 28.9 & 29.0 & 25.2 && \underline{4.4} & 4.3 & \underline{0.5} \\
    ARM & 37.9 & 42.3 & \underline{41.5} && — & — & — && — & — & — && — & — & — \\
    \rowdash
    CORE-T & \textbf{38.6} & \textbf{43.0} & \textbf{41.6} && \textbf{56.9} & \textbf{66.7} & \textbf{66.2} && \textbf{35.8} & \textbf{35.8} & \textbf{33.3} && \textbf{5.3} & \textbf{5.7} & \textbf{1.9} \\
    \rowdash
    Oracle & 47.8 & 50.7 & 50.7 && 64.5 & 71.8 & 71.8 && 65.8 & 65.7 & 65.7 && 6.8 & 7.2 & 7.2 \\
    \addlinespace[4pt]
    \addlinespace[0.5pt]
    \bottomrule
    \end{tabular}
    \end{adjustbox}
    \caption{\textbf{End-to-end SQL execution performance with Llama-3.1-8B-Instruct as table selector.}
    Execution accuracy (EX) on multi-table queries (EX$_{MT}$), all queries (EX$_{all}$), and the perfect-recall subset (EX$_{PR}$), where all gold SQL tables are retrieved and passed to the generator. Oracle uses gold tables; \textbf{best}/\underline{second best} are among non-oracle methods within each SQL-generator block.}
    \label{tab:answer-gen-table}
\end{table*}

\paragraph{How can we improve precision in open-book multi-table retrieval while preserving high recall?}
Table~\ref{tab:arm-db-id-compact} shows that \textsc{CORE-T} improves the precision--PR trade-off over strong baselines by combining compact schema selection with compatibility-based recovery. On \textsc{Bird}, where direct comparison with prior join-aware methods is possible, \textsc{CORE-T} improves F1 over \textsc{ARM}, \textsc{JAR}, and \textsc{REAR} with Llama selection (62.3 vs.\ 53.5/50.1/51.0) while returning fewer tables than \textsc{ARM} and \textsc{ReAct} (4.1 vs.\ 5.3/4.8). With Qwen selection, it also improves over \textsc{ARM} on \textsc{Bird} (72.3 vs.\ 68.3 F1), suggesting that the gain is not tied to a single selector LLM. On \textsc{MMQA}, \textsc{CORE-T} outperforms \textsc{ReAct} and \textsc{REAR} with Qwen selection (56.4 vs.\ 50.4/39.2 F1), indicating that compatibility-guided table-set construction helps when many queries require multi-table reasoning. On the noisier \textsc{Beaver} benchmark, \textsc{CORE-T} achieves the strongest F1 with Qwen (43.0). Although absolute PR is low for every method on \textsc{Beaver}, \textsc{CORE-T} obtains the highest PR with both selectors: 17.7 with Qwen and 15.3 with Llama. Thus, anonymized enterprise-style schemas and empty tables make complete gold-table recovery difficult, exposing an unresolved absolute-performance gap despite \textsc{CORE-T}'s relative improvement. Overall, \textsc{CORE-T} improves precision and F1 while preserving competitive PR, especially when join-coherent table-set construction matters more than single-table relevance.

\paragraph{Additional analyses of threshold stability, prompt length, and table recovery.}
Appendix~\ref{app:table-selection-analysis} presents four supporting analyses showing that the gains are not driven by a brittle threshold or longer prompts. First, sweeping $\tau_{\text{comp}}\in\{0.3,0.5,0.7\}$ changes F1 by at most 4.1 points, PR by at most 1.4 points, and the average number of returned tables by at most 0.4. These results indicate that the threshold provides stable precision--recall control rather than serving as a highly tuned heuristic. We therefore adopt $\tau_{\text{comp}}{=}0.5$ as a shared default across all datasets and selectors. Lower thresholds favor recall, whereas higher thresholds favor compactness (Table~\ref{tab:tau-comp-sweep}). Second, prompt length shows some sensitivity, especially on \textsc{Beaver}, but is not a monotonic explanation of selection quality, suggesting that schema ambiguity, missing values, and weak table semantics also drive errors (Tables~\ref{tab:promptlen-stats} and \ref{tab:promptlen-bins}). Third, adjustment recovers at least one dropped gold table in 9.1--57.1\% of dropped cases and all dropped gold tables in 9.1--50.8\%, confirming its role as a targeted repair step rather than generic expansion (Table~\ref{tab:gold-recovery}). Finally, for low-relevance join-critical bridge tables, adjustment fully recovers dropped bridge tables in 22.5--57.1\% of affected queries, supporting the value of table--table compatibility when query--table relevance is weak (Table~\ref{tab:bridge-recovery}).

\begin{table*}[!tbp]
    \centering
    \small
    \setlength{\tabcolsep}{3pt}
    \renewcommand{\arraystretch}{1.15}
    \resizebox{\textwidth}{!}{%
    \begin{tabular}{@{} l
      r r r
      >{\centering\arraybackslash}m{6pt}
      r r r
      >{\centering\arraybackslash}m{6pt}
      r r r
      >{\centering\arraybackslash}m{6pt}
      r r r
      @{}}
    & \multicolumn{3}{c}{\textbf{Bird}} && \multicolumn{3}{c}{\textbf{Spider}} && \multicolumn{3}{c}{\textbf{MMQA}} && \multicolumn{3}{c}{\textbf{Beaver}} \\
    \cmidrule(lr){2-4} \cmidrule(lr){6-8} \cmidrule(lr){10-12} \cmidrule(lr){14-16}
    & \makecell{\#in tok.\\(M)$\downarrow$} & \makecell{\#out tok.\\(M)$\downarrow$} & \makecell{\#total tok.\\(M)$\downarrow$}
    && \makecell{\#in tok.\\(M)$\downarrow$} & \makecell{\#out tok.\\(M)$\downarrow$} & \makecell{\#total tok.\\(M)$\downarrow$}
    && \makecell{\#in tok.\\(M)$\downarrow$} & \makecell{\#out tok.\\(M)$\downarrow$} & \makecell{\#total tok.\\(M)$\downarrow$}
    && \makecell{\#in tok.\\(M)$\downarrow$} & \makecell{\#out tok.\\(M)$\downarrow$} & \makecell{\#total tok.\\(M)$\downarrow$} \\
    \midrule
    ARM & \cellcolor{red!35} 51.7{\scriptsize\,(4.79$\times$)} & \cellcolor{green!35} 0.73{\scriptsize\,(0.43$\times$)} & \cellcolor{red!35} 52.4{\scriptsize\,(4.20$\times$)} && \multicolumn{3}{c}{---} && \multicolumn{3}{c}{---} && \multicolumn{3}{c}{---} \\
    ReAct & \cellcolor{red!35} 43.5{\scriptsize\,(4.03$\times$)} & \cellcolor{green!35} 1.07{\scriptsize\,(0.63$\times$)} & \cellcolor{orange!25} 44.5{\scriptsize\,(3.57$\times$)} && \cellcolor{red!35} 24.2{\scriptsize\,(4.99$\times$)} & \cellcolor{green!35} 0.74{\scriptsize\,(0.65$\times$)} & \cellcolor{red!35} 24.9{\scriptsize\,(4.16$\times$)} && \cellcolor{red!35} 26.6{\scriptsize\,(4.65$\times$)} & \cellcolor{green!35} 0.84{\scriptsize\,(0.68$\times$)} & \cellcolor{orange!25} 27.5{\scriptsize\,(3.95$\times$)} && \cellcolor{orange!25} 2.6{\scriptsize\,(1.25$\times$)} & \cellcolor{red!35} 1.21{\scriptsize\,(5.05$\times$)} & \cellcolor{orange!25} 3.83{\scriptsize\,(1.64$\times$)} \\
    \rowdash
    CORE-T & \cellcolor{green!35} 10.8{\scriptsize\,(1.0$\times$)} & \cellcolor{green!35} 1.71{\scriptsize\,(1.0$\times$)} & \cellcolor{green!35} 12.5{\scriptsize\,(1.0$\times$)} && \cellcolor{green!35} 4.8{\scriptsize\,(1.0$\times$)} & \cellcolor{green!35} 1.14{\scriptsize\,(1.0$\times$)} & \cellcolor{green!35} 5.99{\scriptsize\,(1.0$\times$)} && \cellcolor{green!35} 5.7{\scriptsize\,(1.0$\times$)} & \cellcolor{green!35} 1.23{\scriptsize\,(1.0$\times$)} & \cellcolor{green!35} 6.96{\scriptsize\,(1.0$\times$)} && \cellcolor{green!35} 2.1{\scriptsize\,(1.0$\times$)} & \cellcolor{green!35} 0.24{\scriptsize\,(1.0$\times$)} & \cellcolor{green!35} 2.33{\scriptsize\,(1.0$\times$)} \\
    \bottomrule
    \end{tabular}%
    }
    \caption{\textbf{Efficiency comparison.}
    Selection-step input/output/total tokens in millions (M), using UAE-Large-V1 embeddings and Llama-3.1-8B-Instruct as selector. Parentheses show factors vs.\ \textsc{CORE-T} in each dataset/metric.}
    \label{tab:efficiency-heat-llama}
\end{table*}

\paragraph{How does table-selection quality impact downstream SQL execution accuracy, especially for multi-table queries?}
Table~\ref{tab:answer-gen-table} shows that better table selection improves downstream EX most clearly on EX$_{MT}$ and for smaller SQL generators, which are less able to ignore distractors or infer missing joins. With Llama-3.2-3B, \textsc{CORE-T} is the strongest non-oracle method on \textsc{Bird}, \textsc{Spider}, and \textsc{MMQA} (15.7, 34.4, and 18.9 EX$_{MT}$), with particularly clear gains over \textsc{REAR} on \textsc{Spider} and \textsc{MMQA} (+8.0 and +10.5 EX$_{MT}$ points). The EX$_{PR}$ results further show that the benefit is not only from recovering all gold tables: on \textsc{MMQA}, \textsc{CORE-T} improves EX$_{PR}$ over \textsc{REAR} by 8.9 points with Llama-3.2-3B (15.8 vs.\ 6.9), indicating that compactness and coherence of the selected schema still matter even when the required tables are available. With stronger generators, gains are smaller but remain visible; with GPT-4o-mini, \textsc{CORE-T} reaches 38.6, 56.9, 35.8, and 5.3 EX$_{MT}$ on \textsc{Bird}, \textsc{Spider}, \textsc{MMQA}, and \textsc{Beaver}. The \emph{Oracle} setting shows substantial headroom (e.g., \textsc{MMQA}: 35.8$\rightarrow$65.8 EX$_{MT}$ with GPT-4o-mini), confirming that table selection remains a bottleneck. \textsc{Beaver} is a stress test: \textsc{CORE-T} achieves the strongest relative F1 and PR, but absolute EX remains low; even the gold-table \emph{Oracle} reaches only 6.8 EX$_{MT}$. This indicates bottlenecks beyond retrieval, including schema anonymization and empty tables. In addition, 20.1\% of \textsc{Beaver}'s gold SQL executions return empty results, making execution scores less stable.

\paragraph{Effect of the table selector (Qwen vs.\ Llama) on downstream EX.}
Table~\ref{tab:answer-gen-table-qwen} (Appendix) repeats the evaluation with Qwen-2.5-7B-Instruct as the selector and preserves the same conclusion: better table selection most reliably improves downstream EX, especially on multi-table queries.

\paragraph{How does our method improve the efficiency of open-book multi-table retrieval?}
Table~\ref{tab:efficiency-heat-llama} reports table-selection LLM token usage for LLM-based selection methods; \textsc{DR@5}, \textsc{JAR}, and \textsc{REAR} are omitted because they do not use an LLM at selection time. Across datasets, \textsc{CORE-T}'s single-shot selector uses fewer input tokens than \textsc{ReAct} (up to 4.99$\times$ lower) and fewer total selection tokens (up to 4.16$\times$ lower). Savings are largest on \textsc{Bird}, \textsc{Spider}, and \textsc{MMQA}, while \textsc{Beaver} shows smaller input-token gains but still lower usage. On \textsc{Bird}, \textsc{CORE-T} also reduces total tokens by 4.20$\times$ relative to \textsc{ARM} (52.4M$\rightarrow$12.5M), reflecting ARM's multi-draft LLM overhead. This matters in open-book retrieval because selection must be repeated for every query over large pooled corpora, making single-shot selection more scalable than iterative or multi-draft LLM pipelines. Overall, \textsc{CORE-T} achieves join-aware behavior with one LLM call, avoiding the iterative overhead of agentic baselines and ARM's combination of MIP-based selection with multiple LLM drafts.

\paragraph{Error analysis.}
We assess query-level significance using a two-sided paired sign-flip test with 10{,}000 randomizations for continuous F1 and exact McNemar tests for binary paired EX, with $p<.05$ considered significant. Across 19 dataset--baseline F1 comparisons, \textsc{CORE-T} is significantly better in 17, tied in one, and worse only on \textsc{Spider} against \textsc{ReAct}. Across 57 EX comparisons over four datasets and three SQL generators, it is significantly better in 30 and tied in 27, with no significant losses (Tables~\ref{tab:paired-f1-all} and~\ref{tab:paired-ex-all}).

Because only \textsc{Bird} provides per-query difficulty labels, we also test seven strata: three difficulty levels and four gold-table-count groups (1, 2, 3, 4+), applying Holm correction separately to F1 and EX. With the Qwen selector and Llama-3.2-3B generator, gains over DR@5 and JAR@5 remain significant in all seven F1 strata and six of seven EX strata, showing that they are not limited to easy or low-table-count queries (Tables~\ref{tab:bird-stratified-f1} and~\ref{tab:bird-stratified-ex}).

The qualitative cases in Table~\ref{tab:qualitative-adjustment-cases} illustrate Adjustment's trade-offs: it can fully restore a missing bridge table (50.0\%$\rightarrow$80.0\% F1), partially restore a join path while adding noise (57.1\%$\rightarrow$66.7\%), or add an unnecessary table when all gold tables are already present (66.7\%$\rightarrow$57.1\%).

Finally, Table~\ref{tab:error-analysis-bird} reports an automated heuristic analysis on \textsc{Bird}, comparing \textsc{CORE-T} with \textsc{ARM} under the same representative setting. \textsc{CORE-T} reduces distractor-table precision errors from 72.5\% (1{,}112 queries) to 55.7\% (855), with a modest increase in recall issues (22.6\% vs.\ 19.0\%). When the generated SQL uses exactly the gold tables, \textsc{CORE-T} nearly eliminates formatting errors (0.1\% vs.\ 7.4\%) but has more schema-linking errors (7.0\% vs.\ 2.4\%). Appendix~\ref{app:error-analysis} provides test details and heuristic definitions.

\section{Ablation Studies}

\begin{table*}[t]
    \centering
    \small
    \setlength{\tabcolsep}{3pt}
    \renewcommand{\arraystretch}{1.1}
    \begin{adjustbox}{max width=\textwidth}
    \begin{tabular}{l l
      c c c
      c c c
      c c c}
    \toprule
    \textbf{Selector} & \textbf{Dataset} &
    \multicolumn{3}{c}{\textbf{DR@10}} &
    \multicolumn{3}{c}{\textbf{+Selection}} &
    \multicolumn{3}{c}{\textbf{+Adjustment (Full)}} \\
    \cmidrule(lr){3-5}\cmidrule(lr){6-8}\cmidrule(lr){9-11}
    & &
    \textbf{F1 (\%)} & \textbf{PR (\%)} & \makecell{\textbf{Avg.\#tab.}$\downarrow$} &
    \textbf{F1 (\%)} & \textbf{PR (\%)} & \makecell{\textbf{Avg.\#tab.}$\downarrow$} &
    \textbf{F1 (\%)} & \textbf{PR (\%)} & \makecell{\textbf{Avg.\#tab.}$\downarrow$} \\
    \midrule
    
    \multirow{4}{*}{\textbf{Qwen-2.5-7B}} &
    \textsc{Bird}   & 31.0 & 94.3 & 10.0 & 80.0 & 79.0 & 2.4 & 72.3 & 87.0 & 3.1 \\
    & \textsc{Spider} & 25.7 & 99.3 & 10.0 & 81.1 & 90.8 & 2.1 & 66.8 & 94.9 & 3.0 \\
    & \textsc{MMQA}   & 29.0 & 66.7 & 10.0 & 63.3 & 51.3 & 2.9 & 56.4 & 59.4 & 4.1 \\
    & \textsc{Beaver} & 33.3 & 25.8 & 10.0 & 43.6 & 15.3 & 4.6 & 43.0 & 17.7 & 5.3 \\
    \midrule
    
    \multirow{4}{*}{\textbf{Llama-3.1-8B}} &
    \textsc{Bird}   & 31.0 & 94.3 & 10.0 & 66.3 & 85.2 & 3.5 & 62.3 & 90.0 & 4.1 \\
    & \textsc{Spider} & 25.7 & 99.1 & 10.0 & 59.7 & 94.4 & 3.5 & 53.8 & 96.6 & 4.2 \\
    & \textsc{MMQA}   & 28.7 & 66.2 & 10.0 & 54.9 & 58.2 & 4.1 & 49.2 & 61.3 & 5.1 \\
    & \textsc{Beaver} & 32.7 & 26.3 & 10.0 & 39.3 & 14.4 & 4.9 & 39.5 & 15.3 & 5.0 \\
    \bottomrule
    \end{tabular}
    \end{adjustbox}
    
    \caption{\textbf{Step-wise ablation of table-set retrieval.}
    F1, perfect recall (PR), and average returned tables for DR@10, after single-shot selection (+Selection), and after additive adjustment (+Adjustment) (full pipeline, $\tau_{\text{comp}}{=}0.5$) on \textsc{Bird}, \textsc{Spider}, \textsc{MMQA}, and \textsc{Beaver} using UAE-Large-V1 as the embedding model.}
    \label{tab:ablation-f1pr-trajectory}
\end{table*}

\paragraph{What does each stage of \textsc{CORE-T} contribute regarding table-set retrieval?}
Table~\ref{tab:ablation-f1pr-trajectory} reports a step-wise ablation (DR@10 $\rightarrow$ +Selection $\rightarrow$ +Adjustment). Selection is the primary driver of compactness and F1, reducing the average number of retrieved tables from 10 to 2.1--4.9 while improving F1 across datasets and selectors. For example, F1 increases from 31.0 to 80.0 on \textsc{Bird} with Qwen and from 25.7 to 59.7 on \textsc{Spider} with Llama, and from 33.3 to 43.6 on \textsc{Beaver} with Qwen. Adjustment adds highly compatible tables to increase PR while minimizing F1 loss. On \textsc{Beaver} with Qwen, F1 decreases by only 0.6 points (43.6$\rightarrow$43.0), while PR increases by 2.4 points (15.3$\rightarrow$17.7); on \textsc{MMQA} with Qwen, PR increases by 8.1 points (51.3$\rightarrow$59.4). Because a missing required table can make correct multi-table SQL impossible, the practical criterion is whether these recall gains improve downstream EX. Selection-only and higher $\tau_{\mathrm{comp}}$ values favor compactness and F1, whereas full \textsc{CORE-T} and lower $\tau_{\mathrm{comp}}$ values favor complete schema coverage. The appropriate choice depends on the downstream generator's sensitivity to missing tables and additional distractors. Appendix~\ref{par:ablation-retrieval-tradeoff} provides further analysis of this recall--precision trade-off.

\begin{table}[t]
    \centering
    \scriptsize
    \setlength{\tabcolsep}{2pt}
    \renewcommand{\arraystretch}{0.95}
    \resizebox{\columnwidth}{!}{%
    \begin{tabular}{l c c}
    \toprule
    \textbf{Stage comparison} & \textbf{EX$_{MT}$} & \textbf{EX$_{all}$} \\
    \midrule
    DR@10 $\rightarrow$ +Selection  & 20 / 0 / 4 & 19 / 2 / 3 \\
    +Selection $\rightarrow$ +Adjustment & 12 / 4 / 8 & 10 / 3 / 11 \\
    DR@10 $\rightarrow$ +Adjustment & 21 / 1 / 2 & 22 / 1 / 1 \\
    \bottomrule
    \end{tabular}
    }
    \caption{\textbf{Execution ablation summary.}
    Entries report the numbers of settings with improved, tied, or reduced execution accuracy across the 24 selector--generator--dataset settings. Full trajectories and detailed results appear in Appendix Table~\ref{tab:ablation-ex-trajectory-combined}.}
    \label{tab:ablation-ex-summary}
\end{table}

\paragraph{Does the full \textsc{CORE-T} pipeline improve downstream SQL execution compared to DR alone?}
Table~\ref{tab:ablation-ex-summary} summarizes the stage-wise execution effects across 24 selector--generator--dataset settings. Selection drives most gains, increasing EX$_{MT}$ in 20 of 24 settings and EX$_{all}$ in 19 of 24. Adjustment is a gated, recall-oriented repair mechanism, not a guarantee of higher EX in every setting: relative to Selection-only, it improves or ties EX$_{MT}$ in 16 of 24 settings and EX$_{all}$ in 13 of 24, but can reduce accuracy when added tables introduce noise. Overall, the full pipeline improves or ties DR@10 in 22 of 24 settings for EX$_{MT}$ and 23 of 24 for EX$_{all}$. Appendix Figures~\ref{fig:em-ge2t-ablation} and~\ref{fig:em-all-ablation} illustrate visually that smaller generators benefit most, while GPT-4o-mini is less sensitive to schema selection. Appendix Table~\ref{tab:ablation-ex-trajectory-combined} reports the full trajectories; the largest gain is 26.3 EX$_{MT}$ points on \textsc{Spider} with Llama-3.2-3B as the SQL generator and Llama-3.1-8B as the selector (8.1$\rightarrow$34.4). These results support our claim that more precise, join-coherent schemas improve execution, especially for weaker SQL generators and multi-table queries. Full \textsc{CORE-T} is preferable when missing join context is costly and excluded tables have strong compatibility with the selected group, whereas Selection-only is preferable when compactness is more important or compatibility signals are noisy.

\section{Conclusion}

We introduced \textsc{CORE-T}, a scalable, training-free framework for \emph{open-book} multi-table retrieval in text-to-SQL over pooled multi-source tables, where \texttt{db\_id} and gold foreign keys are unavailable. \textsc{CORE-T} shifts schema understanding offline via LLM-generated table \emph{purposes} and a lightweight compatibility cache; online, it retrieves top-$K$ candidates, performs a \emph{single} LLM selection guided by relevance and join evidence, and applies a small additive restoration step. Across \textsc{Bird}, \textsc{Spider}, \textsc{MMQA}, and \textsc{BEAVER}, \textsc{CORE-T} returns smaller, more coherent schemas that improve execution, especially on multi-table queries, while reducing token usage. Error analysis shows that \textsc{CORE-T} substantially reduces distractor-table precision errors and nearly eliminates SQL formatting errors, though recall and schema-linking errors remain important areas for improvement. Overall, coherent multi-table retrieval is key for accurate, cost-effective open-book text-to-SQL. Future work includes richer join modeling (e.g., \emph{multi-column}) and extending retrieval to enterprise artifacts (e.g., text and images) via \emph{cross-modal connectivity}.

\section*{Limitations}
Our evaluation is limited by the scope of available benchmarks. We report results
on \textsc{Bird}, \textsc{Spider}, \textsc{MMQA}, and \textsc{Beaver}. For
\textsc{Bird}, \textsc{Spider}, and \textsc{MMQA}, we approximate enterprise
analytics over integrated data sources by merging tables across databases (or
question-specific schemas) into a single pooled corpus and removing
\texttt{db\_ids}. \textsc{Beaver} complements these constructed open-book
settings because it is already released in an open-book form and contains more
realistic, noisy enterprise schemas with anonymized or missing values. However,
\textsc{Beaver} also illustrates the difficulty of evaluating such settings:
many tables are empty due to anonymization, and 20.1\% of gold SQL queries
return empty execution results, which can make downstream EX scores less stable.
Even so, all benchmarks remain simplified relative to real deployments, where
schemas may evolve, access may be governed by organizational constraints, and
data quality issues may be more severe. Moreover, all evaluated benchmarks are
English-only, so we do not assess multilingual open-book retrieval or
text-to-SQL. For \textsc{MMQA}, we only evaluate on a stratified one-third subset
for cost reasons.

Our compatibility cache focuses on key--foreign-key-like joins using column
semantics, value overlap, and simple relational constraints. It may miss other
connections common in practice, including non-equi joins, self-joins, and many-to-many joins via bridge tables. In addition, value-based signals depend on the availability and quality of column values; performance may degrade when values are sparse, heavily skewed, or
unavailable due to privacy constraints. Strict uniqueness and
subset-style constraints can also be brittle under dirty data: duplicates may
violate uniqueness, missing or anonymized values may weaken containment
evidence, and sparse columns may make valid joins hard to detect. Thus, the
cache provides lightweight join evidence, but it should not be interpreted as a
complete model of relational connectivity in arbitrary enterprise data lakes.

Although \textsc{CORE-T} is training-free and uses a single LLM selection call,
it can still be brittle for ambiguous questions or noisy schemas; we mitigate
parsing failures with a robust DR@10 fallback, but this does not eliminate
retrieval errors or schema ambiguity. These limitations motivate future work on
(i) broader connectivity signals beyond strict key--foreign-key patterns,
(ii) soft/ratio-based variants of uniqueness and containment checks to better
tolerate duplicates and missingness,
(iii) incremental and adaptive cache maintenance under evolving schemas, and
(iv) more realistic open-book multi-table retrieval benchmarks and evaluations,
including multilingual settings and controlled noise-injection stress tests.

\section*{Ethics Statement}
We evaluate on publicly available benchmarks (\textsc{Bird}, \textsc{Spider},
\textsc{MMQA}, \textsc{BEAVER}) released for research use under their respective licenses. Our pipeline operates on structured
relational tables and questions and does not collect any new user data or infer
personal or demographic attributes. Pooling tables across databases is used to
simulate integrated data sources and does not introduce additional sensitive
information beyond what is contained in the original datasets.

Our goal is to benefit the research community by improving open-book multi-table
retrieval for text-to-SQL and enabling more efficient, reproducible evaluation.
As with retrieval and generation systems, the approach could be misused in real
deployments to surface or combine information without authorization. However, our approach is intended solely for academic research and is not
designed for deployment in surveillance, decision-making, or other high-stakes
settings. Any
practical use should therefore follow standard data-governance practices (access
control, auditing, and privacy safeguards) and undergo appropriate
oversight.

To support reproducibility and transparency, we document dataset splits, prompts,
and decoding settings (temperature 0). Any future public release of our code or artifacts will follow standard open-source practices, including documentation of intended use, limitations, and guidance for responsible deployment, to help mitigate potential misuse. We also aim to reduce environmental
impact by reusing pretrained models and training-free components rather than
training new large models from scratch. We used AI assistance to help refine
writing and improve presentation.

\section*{Acknowledgments}

We thank Leon Engländer and Shivam Sharma for
their constructive feedback and discussion on this
project. This research work has been funded by the German Federal Ministry of Research, Technology and Space and the Hessian Ministry of Higher Education, Research, Science and the Arts within their joint support of the National Research Center for Applied Cybersecurity ATHENE.


\bibliography{custom}

\appendix

\section{Markdown Serialization of Tables}
\label{app:md-serialization}
We serialize each candidate table into Markdown with a header row,
an alignment row, and \emph{five randomly sampled data rows}. The header preserves original column order and names. The serialized snippet is what the retriever/selector sees.

\begin{lstlisting}[style=prompt, caption={Example Markdown serialization with five randomly sampled rows.}, label={lst:md-serialization}]
Table name: satscores
Example table content:
| cds | rtype | sname | dname | cname | enroll12 | NumTstTakr | AvgScrRead | AvgScrMath | AvgScrWrite | NumGE1500 |
|-------:|-------:|-------:|-------:|-------:|-------:|-------:|-------:|-------:|-------:|-------:|
| 1100170000000 | D |  | Alameda County Office of Education | Alameda | 398 | 88 | 418 | 418 | 417 | 14 |
| 1100170109835 | S | FAME Public Charter | Alameda County Office of Education | Alameda | 62 | 17 | 503 | 546 | 505 | 9 |
| 1100170112607 | S | Envision Academy for Arts & Technology | Alameda County Office of Education | Alameda | 75 | 71 | 397 | 387 | 395 | 5 |
| 1100170118489 | S | Aspire California College Preparatory Academy | Alameda County Office of Education | Alameda | 61 | 0 |  |  |  |  |
| 1611190000000 | D |  | Alameda Unified | Alameda | 922 | 544 | 521 | 546 | 519 | 333 |
\end{lstlisting}

\section{Table Enrichment Details}
\label{app:table-enrichment-details}

\paragraph{Why we use compact purpose descriptions instead of structured metadata?}
A more structured metadata output (e.g., compact JSON fields for key entities or columns) could provide additional signals, but would also increase the selector prompt length and token cost. We therefore adopt a short purpose description as a practical trade-off in the current design. This compact purpose metadata helps control selector prompt length and cost, while the selector still receives the full schema, the 5-row snapshot, and compatibility evidence at inference time; thus, the purpose text is not the only signal available for join-coherent selection.

\paragraph{Why we cap the value snapshot at 5 rows?}
We serialize each table for purpose generation/selection as \emph{name + schema + a small value snapshot} to provide lightweight semantic grounding under a fixed token budget. Thus, prior table-retrieval pipelines commonly serialize tables using a small number of rows/cells alongside schema metadata (e.g., ARM serializes table chunks with metadata and rows, while EnrichIndex uses name/columns plus a random row sample and cites this convention of providing schema plus a small value snapshot as standard in prior work) \citep{chen2025enrichindex, chen-etal-2025-retrieve}. The 5-row cap is therefore an efficiency--signal trade-off that exposes representative values while controlling context length for the table retrieval and selection steps, and is not intended to characterize full value distributions. Importantly, compatibility checks are computed from full columns/contents rather than from the 5-row snapshot.

\section{Compatibility Score Details}
\label{app:compat-eval}

\paragraph{Column-pair score.}
For a cross-table column pair $(c,c')$, we compute: (i) a uniqueness indicator
$u(\cdot)\in\{0,1\}$, (ii) a subset indicator $\mathrm{sub}(c,c')\in\{0,1\}$,
(iii) value overlap $\mathrm{jac}(c,c')\in[0,1]$ (Jaccard), and (iv) header
similarity from an exact lexical score $\mathrm{ex}(c,c')$ and an embedding-based
semantic score $\mathrm{sem}(c,c')$.
We combine header similarities as
\[
\mathrm{name}(c,c')=\tfrac{1}{2}\mathrm{sem}(c,c')+\tfrac{1}{2}\mathrm{ex}(c,c').
\]
We only score plausible key--foreign-key pairs by requiring (a) at least one
column is unique and (b) a subset relation holds:

\begin{equation}
\label{eq:valid}
\begin{aligned}
\operatorname{valid}(c,c') \equiv {}&
\bigl[\max\{u(c),u(c')\}=1\bigr] \\
&\land \bigl[\mathrm{sub}(c,c')=1\bigr].
\end{aligned}
\end{equation}

The column-pair compatibility score is then
\begin{equation}
\label{eq:colscore}
\begin{aligned}
s(c,c') = {}& \mathbb{I}\!\bigl[\operatorname{valid}(c,c')\bigr]\cdot \\
& \Bigl(\tfrac{1}{2}\mathrm{jac}(c,c')+\tfrac{1}{2}\mathrm{name}(c,c')\Bigr).
\end{aligned}
\end{equation}

\begin{figure*}[t]
    \centering
    \includegraphics[width=\linewidth]{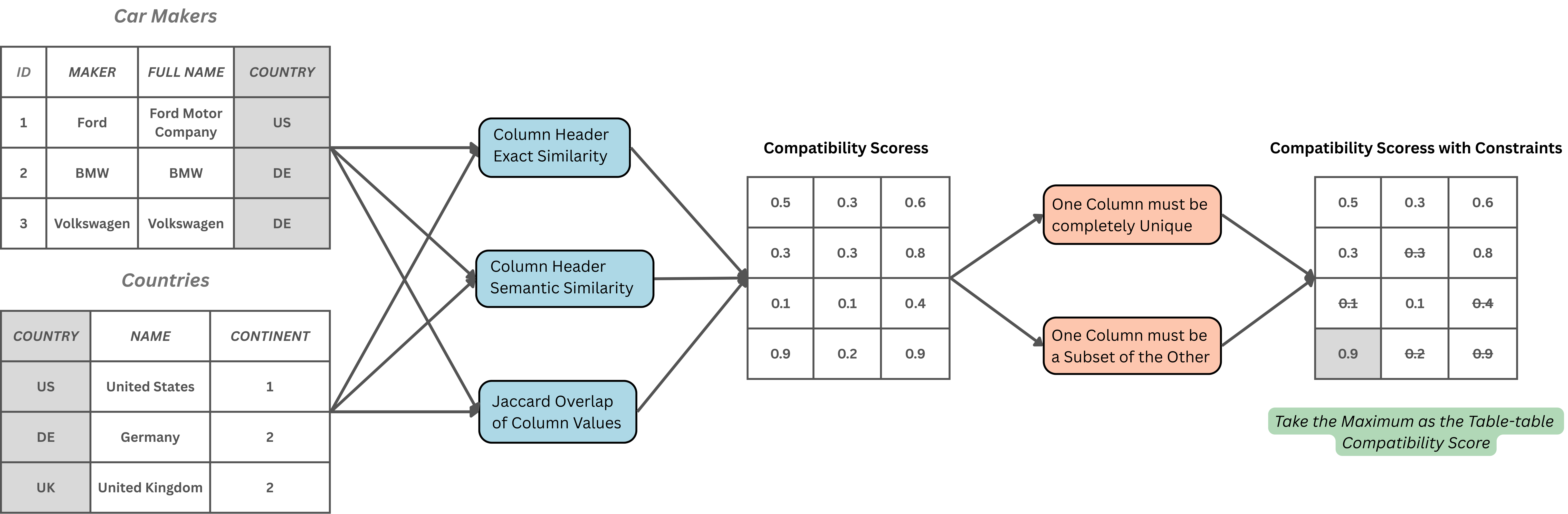}
    \caption{\textbf{Illustration of table-table compatibility scoring.} 
    Given two candidate tables (e.g., \emph{Car Makers} and \emph{Countries}), we compute similarity based on column headers (exact and semantic) and column values (Jaccard overlap). 
    Additional constraints (e.g., uniqueness or subset relations) refine the compatibility assessment. 
    The maximum similarity score under these constraints is taken as the overall table-table compatibility score.}
    \label{fig:compatibility-scores}
\end{figure*}

\paragraph{Table--table score.}
We define table compatibility as the best valid column match:
\begin{equation}
\label{eq:tablescore}
\mathrm{CS}(t_i,t_j)=\max_{c\in C(t_i),\,c'\in C(t_j)} s(c,c'),
\end{equation}
and record the argmax as \texttt{best\_join\_columns}. If no valid pair exists,
$\mathrm{CS}(t_i,t_j)=0$. Figure~\ref{fig:compatibility-scores} illustrates the scoring intuition between two example tables.

\paragraph{Fallback when values are unreliable.}
Dense retrieval is independent of the compatibility cache, and Selection remains operational using the remaining table signals, including names, schemas, sampled values, and generated purpose descriptions, when reliable compatibility evidence is unavailable. In this setting, structural guidance is weaker: missing, sparse, or unreliable values can reduce bridge recovery, but do not disable the CORE-T pipeline. Adjustment therefore refrains from adding tables without sufficient compatibility evidence, preserving query-based retrieval while avoiding speculative compatibility edges and unsupported expansion.

\paragraph{Evaluation protocol.}
We treat a pair of tables as \emph{predicted joinable} if its compatibility score
exceeds \textbf{0.5} ($\mathrm{CS}(t_i,t_j) > 0.5$). Using the gold joinability annotations in
\textbf{BIRD} and \textbf{SPIDER}, we compute:
(i) \textbf{Joinability Accuracy} --- whether our binary prediction
(joinable / not joinable) matches the gold label; and
(ii) \textbf{Column-Pair Accuracy} --- among gold-joinable pairs, whether the
column pair with the highest predicted compatibility score matches the gold
(join) column pair.

For the \textbf{Average Compatibility Score Difference}, we assign a gold value
\(g\in\{0,1\}\) to each table pair (1 if joinable, 0 otherwise), let \(s\in[0,1]\)
be our predicted compatibility score, compute \(|s-g|\) for each pair, and report
the average across pairs (lower is better) (Table~\ref{tab:compatibility-eval}). We do not report these metrics on
\textbf{MMQA} because the dataset does not provide gold table--table joinability
signals.

\paragraph{Comparison to a JAR-style compatibility score.}
To disentangle the effect of our enforced relational constraints, we also evaluate a JAR-inspired
variant that scores column pairs using only header similarity and value overlap (i.e., without
requiring uniqueness/subset validity and without ignoring \textsc{null}s)). Table~\ref{tab:compatibility-eval} shows that enforcing these
constraints in \textsc{CORE-T} improves joinability accuracy and column-pair identification, and reduces the average
compatibility-score error.

\begin{table*}[t]
\centering
\small
\setlength{\tabcolsep}{6pt}
\renewcommand{\arraystretch}{1.15}
\begin{adjustbox}{max width=\textwidth}
\begin{tabular}{l
  c c c
  >{\centering\arraybackslash}m{6pt}
  c c c}
& \multicolumn{3}{c}{\textbf{Bird}} && \multicolumn{3}{c}{\textbf{Spider}} \\
\cmidrule(lr){2-4} \cmidrule(lr){6-8}
& \textbf{Join. Acc.} & \textbf{Col.-Pair Acc.} & \textbf{Avg $|s-g|$}
&& \textbf{Join. Acc.} & \textbf{Col.-Pair Acc.} & \textbf{Avg $|s-g|$} \\
\midrule
JAR
& 77.2\% & 35.2\% & 0.258 && 73.5\% & 41.4\% & 0.293 \\
\rowdash
\textsc{CORE-T}
& \textbf{97.3\%} & \textbf{84.5\%} & \textbf{0.092}
&& \textbf{89.1\%} & \textbf{69.0\%} & \textbf{0.104} \\
\bottomrule
\end{tabular}%
\end{adjustbox}
\caption{\textbf{Compatibility scores evaluation on Bird and Spider.}
We compare a JAR-style compatibility formula scores against \textsc{CORE-T}'s constrained formula scores. We report
(i) joinability accuracy, (ii) column-pair accuracy, and (iii) average score error $|s-g|$
(gold $g\in\{0,1\}$; lower is better). MMQA is excluded due to missing gold joinability annotations.}
\label{tab:compatibility-eval}
\end{table*}

\paragraph{Scalability and incremental maintenance.}
Our scalability claim targets \emph{online} inference: at query time, \textsc{CORE-T} does not perform corpus-wide all-pairs compatibility checks. A naïve eager materialization of $\mathrm{CS}(\cdot,\cdot)$ for all table pairs in a corpus of $N$ tables would be $O(N^2)$ table-pair evaluations (and can be parallelized/sharded offline). However, a full $N^2$ cache is not required for retrieval-time use: the selection/adjustment steps only need compatibilities among a small candidate set (e.g., DR top-$K$ tables). Thus, for large corpora, the cache can be constructed \emph{sparsely} by computing and storing entries only for candidate pairs encountered during inference, yielding $O(K^2)$ evaluated pairs per query (and $O(Q\cdot K^2)$ over $Q$ queries), while enabling fast reuse once computed.

This sparse view also supports \emph{dynamic updates}. When a new table arrives or an existing table changes, we only need to compute compatibilities involving that table (i.e., against other candidate tables or against the current corpus), rather than rebuilding the entire cache. Concretely, entries can be computed on demand (``lazy caching'') for candidate pairs and stored so future queries reuse the result.

\section{CORE-T Pipeline Details}
\label{app:core-t-pipeline-details}

\subsection{Table Selection Details}
\label{app:table-selection}

\begin{figure*}[t]
  \centering
  \includegraphics[width=\textwidth]{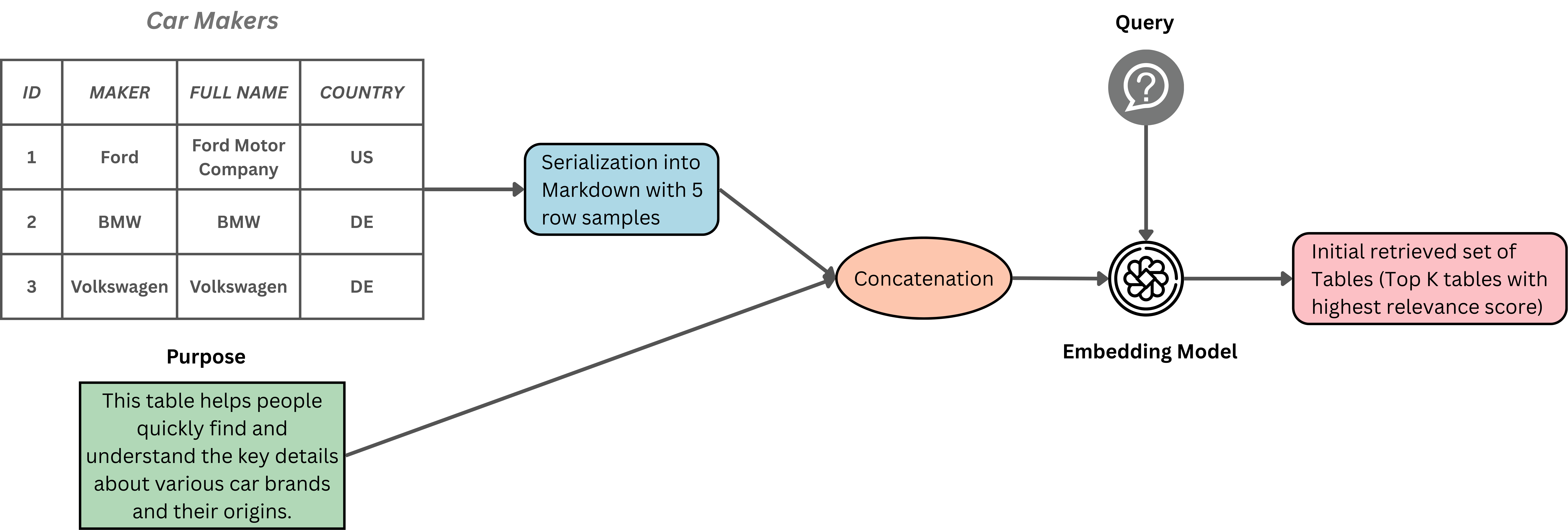}
  \caption{\textbf{Initial dense retrieval with enriched table embeddings.}
  Offline, each table is serialized into Markdown (with 5-row samples) and augmented with an LLM-generated purpose description; the concatenated text is embedded and indexed.
  At inference time, the query is embedded in the same space and the top-$K$ tables are retrieved by cosine similarity, forming the high-recall candidate set $T_K(q)$.}
  \label{fig:initial-table-retrieval}
\end{figure*}

\paragraph{Prompted behavior.}
The LLM is guided through a fixed reasoning policy. We tried to replicate a policy with detailed instructions oriented at the human reasoning workflow:

\begin{enumerate}
    \item \textbf{Understand the query.} Identify core entities and relationships, and what type of data is required to answer the query ($q$).
    \item \textbf{Evaluate individual table relevance.} Use table names, column headers, and sample rows to judge whether each table is relevant. When unsure, the model is explicitly instructed to treat a table as \emph{potentially} relevant instead of discarding it.
    \item \textbf{Evaluate pairwise compatibility.} For each pair of retrieved tables with compatibility analysis, interpret the $\mathrm{CS}$ scores and best join columns, cross-checking with column names and sample values. Again, when in doubt, the model is instructed to treat the pair as potentially joinable.
    \item \textbf{Group formation.} Form one or more groups of tables where all members are joinable, i.e., groups that form connected join graphs under the provided compatibility edges. The model is encouraged to prefer larger groups when there is uncertainty, rather than splitting aggressively.
    \item \textbf{Group selection.} Select a \emph{single} most relevant and compatible group for answering the query, emphasizing high recall: tables that are plausibly useful should be retained to avoid missing necessary information.
\end{enumerate}

The model is further instructed \emph{not} to aggressively eliminate tables and to only remove a table when it is clearly irrelevant or incompatible.

\paragraph{Output and selected subset.}
The LLM returns a JSON object that includes:
\begin{itemize}
    \item a list of formed groups, each with a \texttt{group\_index} and its member \texttt{table\_indices};
    \item a \texttt{selected\_group\_index} indicating which group should be used to answer $q$.
\end{itemize}
We also allow the model to output textual rationales before selecting tables for better reasoning and debugging, but ignore them at result extraction. We parse the JSON and take the tables belonging to the selected group as the LLM-selected subset.

\subsection{Table Adjustment Details}
\label{app:table-adjustment}

\paragraph{Greedy expansion.}
Starting from the nonempty augmented seed group $G_0 \subseteq T_K(q)$, we initialize
$G\leftarrow G_0$ and greedily consider tables from $T_K(q)\setminus G$ until
no candidate yields positive marginal gain.
Let $\mathrm{CS}(t_i,t_j)\in[0,1]$ denote the cached pairwise compatibility score between
tables $t_i,t_j\in T_K(q)$. We define the group's mean pairwise compatibility and
mean retriever relevance as
\begin{small}
\begin{align}
\overline{\mathrm{CS}}(G)
&=
\begin{cases}
\dfrac{1}{\binom{|G|}{2}}
\sum_{\substack{t_i,t_j\in G\\ i<j}} \mathrm{CS}(t_i,t_j), & |G|>1,\\
0, & |G|\le 1,
\end{cases}
\label{eq:mean-compat}\\[-1mm]
\overline{\mathrm{RS}}(q,G)
&=\frac{1}{|G|}\sum_{t\in G}\mathrm{RS}(q,t),
\label{eq:mean-rs}
\end{align}
\end{small}
where $\mathrm{RS}(q,t)$ is the dense-retrieval relevance score defined in the
initial retrieval stage. The eligible candidate set contains tables with nonzero
compatibility to at least one current group member:
\begin{equation}
\mathcal{E}(G)=\left\{t\in T_K(q)\setminus G:
\max_{t'\in G}\mathrm{CS}(t,t')>0\right\}.
\label{eq:edge-eligibility}
\end{equation}
Among eligible candidates, we select the table with the largest marginal gain:
\begin{align}
\Delta(t;G)
&= w_{\mathrm{coh}}\Delta_{\mathrm{CS}}(t;G)
 + w_{\mathrm{rel}}\Delta_{\mathrm{RS}}(t;G), \label{eq:marginal-gain}\\[-1mm]
\Delta_{\mathrm{CS}}(t;G)
&= \overline{\mathrm{CS}}(G\cup\{t\})-\overline{\mathrm{CS}}(G),\notag\\
\Delta_{\mathrm{RS}}(t;G)
&= \overline{\mathrm{RS}}(q,G\cup\{t\})-\overline{\mathrm{RS}}(q,G).\notag
\end{align}
We set $w_{\mathrm{coh}}=w_{\mathrm{rel}}=\tfrac{1}{2}$ in our experiments.
If $\mathcal{E}(G)$ is empty, expansion stops. Otherwise, we set
$t^\star=\arg\max_{t\in\mathcal{E}(G)}\Delta(t;G)$ and add it only if
$\Delta(t^\star;G)>0$; we then update $G\leftarrow G\cup\{t^\star\}$ and repeat.
The final schema $S(q)=G$ is a compact
table group that preserves the LLM-selected seed tables while recovering additional
compatible and relevant tables for SQL generation.


\section{Dataset Preprocessing Details}
\label{app:dataset-preprocessing}

\paragraph{\textsc{Bird} and \textsc{Spider}.}
Both benchmarks are released in a closed-book setting where each query is
associated with a known database identifier (\texttt{db\_id}). To instantiate our
open-book setting, we merge all databases in the dev split into a single pooled
table corpus per benchmark and drop \texttt{db\_id}. Retrieval is then performed
over the entire pooled corpus rather than within a pre-selected schema.

\paragraph{\textsc{MMQA}.}
In the original \textsc{MMQA} setup, each question is paired with a small
question-specific set of tables. We convert it to an open-book corpus by merging
all tables across questions into one pool and renaming tables with conflicting
names but different schemas so that each table has a unique name. The pooled
corpus contains 710 tables and 3{,}313 questions.

\paragraph{Stratified sampling for \textsc{MMQA}.}
To reduce computational cost and keep the evaluation size comparable to
\textsc{Bird} and \textsc{Spider}, we stratify \textsc{MMQA} questions by the
number of gold tables and sample one third from each stratum, yielding 1{,}105
queries. Table~\ref{tab:dataset-stats} reports statistics for this evaluation subset.

\paragraph{\textsc{BEAVER}.}
\textsc{BEAVER} is an enterprise text-to-SQL benchmark constructed from
anonymized subsets of real-world enterprise data warehouses. The dataset covers university facilities data, enterprise networking and virtual-machine infrastructure. Unlike \textsc{Bird}, \textsc{Spider}, and \textsc{MMQA}, \textsc{BEAVER} is already released in a form suitable for open-book retrieval, so no additional pooling or schema preprocessing is needed beyond removing the \texttt{db\_id}. It is also deliberately challenging for retrieval: many tables are vague, semantically similar, or only weakly distinguishable by surface names and columns, making them easy distractors for retrievers. At the same time, because \textsc{BEAVER} is derived from real enterprise data and passed through an anonymization pipeline, many values were removed. In the original corpus, 214 of 463 SQL tables are completely empty, and a large fraction of gold SQL queries (20.1\%) therefore return empty result sets, which affects our SQL execution evaluation. This makes \textsc{BEAVER} a realistic but particularly noisy and difficult stress test for open-book table retrieval and downstream SQL evaluation.

\paragraph{Additional benchmarks for future evaluation.}
Beyond \textsc{Bird}, \textsc{Spider}, \textsc{MMQA}, and \textsc{BEAVER}, recent enterprise-oriented benchmarks such as
\textsc{Spider 2.0} is a promising target for broader coverage. However, adapting it to our open-book pooled-table formulation
requires additional preprocessing (e.g., pooling tables into a single corpus and removing scoping identifiers when present) and additional infrastructure to handle very large enterprise tables, in practice (notably resource-heavy SQL execution), which we plan to pursue in follow-up experiments.

\section{Baseline Details}
\label{app:baselines}

\paragraph{Summary comparison.}
Table~\ref{tab:efficiency-bird-simple} provides a high-level comparison of the
baselines in terms of qualitative LLM-call cost and whether \texttt{db\_id} is
assumed during retrieval/selection.

\begin{table*}[!tbp]
\centering
\large 
\setlength{\tabcolsep}{3pt}
\renewcommand{\arraystretch}{1.1}
\begin{adjustbox}{max width=\textwidth}
\begin{tabular}{@{} l c c l @{}}
\toprule
\textbf{Method} & \textbf{\makecell{Est.\ LLM \\ calls/query}} & \textbf{\makecell{\texttt{db\_id} in \\ retr./sel.}} & \textbf{Rationale} \\
\midrule
DR    & \makecell{\textbf{None} \\ (0)}
      & \xmark
      & \makecell[l]{Ranks tables by embedding similarity;
                    one short LLM call is done offline\\ per table to generate
                    purpose metadata; no combinatorial search.} \\
\rowdash
ReAct & \makecell{\textbf{High} \\ (no. of iters.)}
      & \xmark
      & \makecell[l]{Cost grows roughly linearly with number
                    of iterations\\ and retrieved items;
                    avoids combinatorial MIP search.} \\
\rowdash
JAR   & \makecell{\textbf{Low} \\ (1 small call)}
      & \cmark
      & \makecell[l]{Uses MIP to re-rank tables and columns, which
                    grows exponentially\\ with more tables/columns.
                    LLM used only for light query decomposition.} \\
\rowdash
ARM   & \makecell{\textbf{High} \\ (1 small+3 big)}
      & \cmark
      & \makecell[l]{Expands $K$ retrieved table sets \& re-ranks them
                    with MIP, causing\\ exponential growth.
                    Drafts multiple LLM candidates for $K{=}2,3,4$.} \\
\rowdash
REAR & \makecell{\textbf{Low} \\ (1 small call)}
      & \xmark
      & \makecell[l]{Uses LLM to generate offline table descriptions for retrieving base tables,\\      expands them with column-embedding, then refines with cross-encoder scoring.} \\
\rowdash
CORE-T  & \makecell{\textbf{Medium} \\ (1 big call)}
      & \xmark
      & \makecell[l]{Performs a single selection pass over top-$K$
                    tables plus light\\ filtering/adjustment.
                    Scales roughly linearly with tables and columns.} \\
\bottomrule
\end{tabular}%
\end{adjustbox}
\caption{\textbf{Rough efficiency comparison of baselines and SOTA methods for table retrievers.}
“Low/Medium/High” indicate qualitative LLM-call cost.
The column “\texttt{db\_id} in retr./sel.?” marks whether a method assumes database identifiers during \emph{table retrieval and selection}: JAR and ARM assume availability (\cmark), while DR, ReAct, REAR, and CORE-T do not (\xmark).}
\label{tab:efficiency-bird-simple}
\end{table*}

\paragraph{Dense Retriever (DR).}
For each table, we generate a short \emph{table purpose} and append it to the
table’s 5-row Markdown serialization (cf. \S\ref{lst:md-serialization}). We
embed this concatenated text (e.g., with UAE-Large-V1) to obtain a table vector.
Given a query, we embed it into the same space and retrieve the top-$K$ tables
by cosine similarity. We denote this baseline as \textbf{DR@$K$}.

\begin{lstlisting}[style=prompt, caption={Example table encoded for dense retrieval.}, label={lst:dense-encoding}]
Table name: satscores
Table purpose: This table appears to be a collection of data about schools in Alameda County, specifically their performance on standardized tests. The table includes information such as the school's name, type, and enrollment, as well as the average scores of their students in reading, math, and writing. It also tracks the number of students who scored at or above a certain threshold (1500) on these tests. This data can be used to compare the performance of different schools and identify areas where they may need improvement.
Table content: {serialized table markdown}
\end{lstlisting}

\paragraph{ReAct.}
ReAct interleaves \emph{Thought}, \emph{Action}, and \emph{Observation} steps.
In each action, the agent calls \texttt{table\_search} with generated keywords;
\texttt{table\_search} queries the same dense index as DR and returns up to 5
\emph{new} tables (deduplicated within a run). We cap ReAct at 3 tool calls per
question and use a recall-oriented prompt (``when in doubt, include the table'').

When the agent stops, it outputs a JSON object containing \texttt{relevant\_tables}
(table indices). We parse this list and treat it as the predicted table set.
If the output is invalid (e.g., empty or unparsable), we fall back to \textbf{DR@$K$}
and use the top-$K$ tables as the prediction.

\paragraph{JAR.}
JAR is a join-aware table-retrieval re-ranker. Given
an initial set of candidates, it infers a join graph and solves a mixed-integer
program (MIP) that selects a \emph{connected} set of $K$ tables by jointly
balancing (i) query coverage/relevance and (ii) table--table join compatibility,
rather than ranking tables independently. In our pooled setting, we run the
authors' public implementation with their default hyperparameters.

\paragraph{ARM.}
ARM is an LLM-guided \emph{retrieve-all-at-once}
retriever for complex table QA (evaluated by the authors on \textsc{BIRD}). It
first performs an \emph{information-alignment} stage to retrieve candidate tables
(e.g., aligned keywords/$n$-grams combined with embedding-based search), and then
runs a join-aware \emph{structure-alignment} stage that uses a MIP (similar in
spirit to JAR) to select a small \emph{connected} table set that jointly
maximizes query--table relevance and table--table compatibility (e.g., through
joinable columns). Finally, ARM applies LLM self-verification/aggregation to finalize the retrieved set (e.g., by generating multiple LLM drafts and aggregating their selected tables' logit scores). In our experiments, we use the authors' released pipeline and default hyperparameters where supported.

\paragraph{REAR.}
REAR is a retrieve--expand--refine framework for multi-table retrieval. It first
retrieves query-relevant base tables using a standard retriever over offline
LLM-generated table descriptions, then expands this set by searching for
structurally joinable tables with precomputed column embeddings and approximate
nearest-neighbor search. The expanded pool is then refined by jointly scoring
query--table relevance and table--table joinability with cross-encoder
reranking. For this baseline, we use the authors' released predicted tables from
their best reported configuration.

\paragraph{Comparison: \textsc{CORE-T} vs.\ JAR/ARM/REAR.}
Although all four methods are join-aware, they differ in \emph{where} joinability enters the pipeline and how
robust the selection step is under open-book noise:

\begin{itemize}
  
  \item \textbf{How joinability is used in the pipeline.}
  JAR and ARM both rely on an explicit join graph/compatibility structure to
  \emph{drive} selection through a connectivity-constrained MIP: the optimizer
  searches for a \emph{connected} table set that optimizes for relevance and
  compatibility scores. ARM additionally performs LLM-guided alignment to
  propose candidates before the MIP selection. REAR instead uses joinability in
  a retrieve--expand--refine pipeline: it expands initially retrieved tables
  with column-embedding-based joinable candidates and then refines the expanded
  pool with query--table and table--table reranking. In contrast,
  \textsc{CORE-T} uses compatibility scores as \emph{evidence} for an LLM
  selector: we condition a single selection pass on table purposes and
  compatibility edges, and then apply a two-step additive restoration step
  that re-inserts strongly compatible tables that were pruned early. Thus,
  \textsc{CORE-T} uses compatibility as structured guidance during selection
  and as a targeted mechanism to obtain a more precise table set while
  protecting recall.

  \item \textbf{Joinability quality under integrated corpora.}
  In open-book collections, spurious matches (e.g., shared column names like
  \texttt{id}, \texttt{name}) can create dense, noisy compatibility structure.
  This can affect MIP-based methods such as JAR and ARM, whose final selection is
  sensitive to the induced join graph. It can also affect REAR, because its
  expansion stage is driven by column-level embedding similarity: semantically
  similar columns across unrelated domains can introduce plausible but incorrect
  expansion candidates. We compare the JAR compatibility formula against our
  scoring function and find that adding simple relational constraints
  (key-likeness/uniqueness and subset containment) substantially improves
  joinability accuracy and join-column identification
  (Appendix~\ref{app:compat-eval}, Table~\ref{tab:compatibility-eval}),
  reducing false join edges that can mislead both connectivity-based selection
  and expansion-based retrieval.

  \item \textbf{Semantic disambiguation.}
  \textsc{CORE-T} uses LLM-generated \emph{table purpose} to distinguish tables
  that are lexically or structurally similar but differ in intent and
  relationships, a common failure mode in integrated corpora (e.g., multiple
  plausible \texttt{buildings} tables from different domains). This is especially
  important when join evidence alone is ambiguous: JAR/ARM may over-trust
  induced compatibility edges, while REAR may expand toward tables with similar
  column semantics but its reliance on corpus-wide column-embedding
  joinability can make it vulnerable to cross-domain near duplicates and
  semantically similar non-joinable columns.

  \item \textbf{Robustness of selection under noise.}
  Because MIP selection is sensitive to the induced join graph, a small number of
  incorrect edges can steer JAR toward a connected but wrong subset. ARM is more
  robust than JAR in this respect because it includes self-aggregation and
  verification: it generates multiple LLM drafts and aggregates/votes to finalize
  the table set, at the cost of additional LLM overhead. REAR avoids this
  overhead and improves recall through join-aware expansion, but its refinement
  remains a reranking/pruning step over candidates produced by local
  column-similarity evidence. \textsc{CORE-T} instead uses a single selection
  pass conditioned on relevance, purpose metadata, and compatibility evidence,
  and then restores highly compatible tables to protect recall, avoiding both
  iterative search and multi-draft overhead.

  \item \textbf{\texttt{db\_id} assumption.}
  Both JAR and ARM assume access to database identifiers (\texttt{db\_id}) during retrieval/selection, which provides
  additional schema-level scoping signals compared to our open-book setting where \texttt{db\_id} is unavailable (as in
  integrated enterprise corpora without a clean separation into databases). REAR is closer to our setting because it does
  not require \texttt{db\_id}.
\end{itemize}

\section{Experimental Setup Details}
\label{app:experimental-setup-details}

\subsection{Evaluation Metrics}
\label{app:eval-metrics}

\paragraph{Retrieval and selection.}
For each query $q$, we compare the predicted table set $S(q)$ to the gold tables
$G(q)$ and report precision, and F1. We additionally report
\emph{perfect recall} (PR), the fraction of queries for which $G(q)\subseteq S(q)$.

\paragraph{End-to-end performance.}
We report execution accuracy (EX), counting a prediction as correct if executing the
generated SQL yields the same result as executing the gold SQL. Since the SQL
generator only observes the selected tables, EX measures the downstream effect of
table selection on final SQL execution. We report EX on all queries (EX$_{all}$),
on multi-table queries involving at least two gold tables (EX$_{MT}$), and on the
perfect-recall subset (EX$_{PR}$), where all gold tables required by the gold SQL
are retrieved and included among the tables passed to the SQL generator. EX$_{PR}$ helps isolate table-retrieval errors from SQL-generation errors by evaluating only cases where the required gold tables are available to the SQL generator.

\paragraph{Efficiency.}
We report table-selection efficiency using LLM input, output, and total token counts. For agentic baselines, counts are summed across iterations.

\subsection{Implementation Details}
\label{app:implementation-details}

\paragraph{SQL generation LLMs.}
For SQL generation, we use GPT\mbox{-}4o\mbox{-}mini via OpenAI's API and also report
results for two open-source models with Llama\mbox{-}3.2\mbox{-}3B and Gemma\mbox{-}3\mbox{-}4B.

\paragraph{SQL execution and timeout.}
We execute the generated SQL against the selected tables with a \textbf{60-second} timeout per query. If execution exceeds this limit (e.g., due to inefficient joins or malformed queries), we treat the prediction as a
\emph{failed} execution and count it as incorrect for execution accuracy (EX) metric.

\paragraph{Hardware environment.}
All experiments are conducted on a single NVIDIA A100 GPU (with \textbf{40\,GB} of VRAM)
and a machine with \textbf{32\,GB} of system RAM. We used the Hugging Face Transformers library~\citep{wolf-etal-2020-transformers} for running LLM inference.

\paragraph{Decoding and prompt configuration.}
For all LLM calls (purpose generation, table selection, ReAct, and SQL generation), we fix the sampling configuration to: temperature $= 0$, top\mbox{-}k sampling with $k = 1$, top\mbox{-}p sampling with $p = 1.0$, and random seed set to 42.

\paragraph{Embeddings, thresholds and table-selection fallback.}
We use UAE\mbox{-}Large\mbox{-}V1 as the default embedding model for initial table retrieval and all dense-retrieval baselines. We use it in its standard embedding mode (no explicit task instruction prefix); our indexed table text is enriched with LLM-generated purpose descriptions, providing lightweight task conditioning for semantic table matching. We set the dense retrieval cutoff to $K{=}10$ and use a fixed adjustment threshold $\tau_{\text{comp}}{=}0.5$ across datasets to balance recall and the number of tables passed to the SQL generator, keeping our pipeline computationally efficient.
These parameters are kept fixed across all datasets.
If our table-selection LLM call fails or its JSON output cannot be parsed to extract the selected tables, we fall back to the DR@10 set $T_K(q)$. In practice, this fallback is rarely triggered (in fewer than 1\% of queries across our runs).

\paragraph{Reproducibility note on ARM beyond supported benchmarks}
In our experiments, we run ARM using the authors’ publicly released pipeline on the only text-to-SQL dataset it currently supports (\textsc{Bird}).
However, extending ARM to additional text-to-SQL benchmarks is not currently straightforward because the released pipeline depends on dataset-specific intermediate artifacts (e.g., pre-computed alignment/similarity scores produced after dataset preprocessing and chunking/splitting decisions) that are only provided for the datasets covered in the original work.
While the paper describes the high-level stages of the method, re-implementing the full artifact-generation pipeline from scratch for new benchmarks is challenging with the lack of implementation details (e.g., precise preprocessing and chunking/splitting strategies or sizes needed to reproduce the same intermediate scores).
As a result, ARM’s public release is not readily reproducible \emph{beyond} its supported datasets at the time of writing, which can act as a practical reproducibility blocker for the community when attempting broader cross-benchmark evaluation.
We therefore report ARM results only on supported datasets (\textsc{Bird}) and encourage future releases to include the missing artifact-generation details or scripts to enable dataset extension and full reproducibility.

\paragraph{Reproducibility note on JAR beyond supported benchmarks.}
JAR provides reproducible scripts for \textsc{Bird} and \textsc{Spider}, but extending it to additional benchmarks is also non-trivial in practice because its MIP re-ranking objective relies on dataset-specific hyperparameters tuned for those supported datasets.
The released code includes tuned settings for \textsc{Bird}/\textsc{Spider}, but the procedure for how these hyperparameters are optimized (e.g., search space, tuning split, objective, and stopping criteria) is not fully specified, making it difficult to reproduce the same tuning process or fairly adapt JAR to new datasets under a consistent protocol.
Accordingly, we report JAR results only on the datasets where the authors provide tuned hyperparameters and runnable scripts.

\begin{table}[!tbp]
    \centering
    \small
    \setlength{\tabcolsep}{5pt}
    \renewcommand{\arraystretch}{1.1}
    \begin{tabular}{l l r r r}
    \toprule
    \textbf{Dataset} & \textbf{Selector} & \textbf{Min} & \textbf{Median} & \textbf{Max} \\
    \midrule
    
    \multirow{2}{*}{\textsc{Bird}}
    & Llama-3.1-8B & 3{,}916 & 6{,}710 & 12{,}860 \\
    & Qwen-2.5-7B   & 4{,}162 & 7{,}762 & 14{,}301 \\
    \midrule
    
    \multirow{2}{*}{\textsc{Spider}}
    & Llama-3.1-8B & 3{,}345 & 4{,}566 & 6{,}559 \\
    & Qwen-2.5-7B   & 3{,}257 & 4{,}667 & 7{,}032 \\
    \midrule
    
    \multirow{2}{*}{\textsc{MMQA}}
    & Llama-3.1-8B & 3{,}399 & 5{,}182 & 7{,}184 \\
    & Qwen-2.5-7B   & 3{,}473 & 5{,}514 & 8{,}463 \\
    \midrule
    
    \multirow{2}{*}{\textsc{Beaver}}
    & Llama-3.1-8B & 3{,}768 & 9{,}305 & 25{,}223 \\
    & Qwen-2.5-7B   & 5{,}244 & 11{,}086 & 27{,}894 \\
    \bottomrule
    \end{tabular}
    
    \caption{\textbf{Selector prompt token statistics.}
    Min/median/max selector input token counts under the fixed 5-row snapshot used for table serialization and selection.}
    \label{tab:promptlen-stats}
\end{table}

\paragraph{Author correspondence for complete comparisons.}
Our goal is to encourage standardized and extensible evaluation that supports fair cross-method comparisons and reduces friction when benchmarking new retrievers on additional datasets, and helps advance open-book multi-table retrieval research. Accordingly and until the time of writing, we tried contacting the main author multiple times to request clarification and the missing materials needed to reliably extend the released JAR/ARM pipelines beyond their supported datasets. AS these were not available, we restrict our comparisons accordingly.



\begin{table*}[!tbp]
    \centering
    \small
    \setlength{\tabcolsep}{3pt}
    \renewcommand{\arraystretch}{1.1}
    \begin{tabular}{l l c c c c c c c c c}
    \toprule
    \textbf{Selector} & \textbf{Dataset} &
    \multicolumn{3}{c}{$\boldsymbol{\tau_{\text{comp}}=0.3}$} &
    \multicolumn{3}{c}{$\boldsymbol{\tau_{\text{comp}}=0.5}$} &
    \multicolumn{3}{c}{$\boldsymbol{\tau_{\text{comp}}=0.7}$} \\
    \cmidrule(lr){3-5}\cmidrule(lr){6-8}\cmidrule(lr){9-11}
    & &
    \textbf{F1} & \textbf{PR} & \makecell{\textbf{Avg.\#tab.}$\downarrow$} &
    \textbf{F1} & \textbf{PR} & \makecell{\textbf{Avg.\#tab.}$\downarrow$} &
    \textbf{F1} & \textbf{PR} & \makecell{\textbf{Avg.\#tab.}$\downarrow$} \\
    \midrule
    
    \multirow{4}{*}{\textbf{Llama-3.1-8B}} &
    \textsc{Bird}   & 61.3\% & 90.5\% & 4.2 & 62.3\% & 90.0\% & 4.1 & 64.2\% & 89.5\% & 3.9 \\
    & \textsc{Spider} & 53.2\% & 96.6\% & 4.3 & 53.8\% & 96.6\% & 4.2 & 55.6\% & 96.3\% & 4.0 \\
    & \textsc{MMQA}   & 49.1\% & 61.3\% & 5.1 & 49.2\% & 61.3\% & 5.1 & 49.9\% & 61.2\% & 4.9 \\
    & \textsc{Beaver} & 39.5\% & 15.3\% & 5.1 & 39.5\% & 15.3\% & 5.0 & 39.6\% & 15.3\% & 5.0 \\
    \midrule
    
    \multirow{4}{*}{\textbf{Qwen-2.5-7B}} &
    \textsc{Bird}   & 71.0\% & 87.2\% & 3.2 & 72.3\% & 87.0\% & 3.1 & 75.1\% & 85.8\% & 2.9 \\
    & \textsc{Spider} & 65.9\% & 94.9\% & 3.0 & 66.8\% & 94.9\% & 3.0 & 68.6\% & 94.1\% & 2.8 \\
    & \textsc{MMQA}   & 56.3\% & 59.6\% & 4.2 & 56.4\% & 59.4\% & 4.1 & 57.3\% & 59.0\% & 4.0 \\
    & \textsc{Beaver} & 42.6\% & 18.2\% & 5.4 & 43.0\% & 17.7\% & 5.3 & 43.2\% & 17.2\% & 5.2 \\
    \bottomrule
    \end{tabular}
    \caption{\textbf{Sensitivity of \textsc{CORE-T} to the compatibility threshold $\tau_{\text{comp}}$.}
    We report F1, perfect recall (PR), and average returned tables (Avg.\#tab.) for the \textsc{CORE-T} full pipeline.
    Sweeping $\tau_{\text{comp}}\in\{0.3,0.5,0.7\}$ produces small changes in F1/PR and average returned tables, indicating that $\tau_{\text{comp}}$ is a stable, interpretable trade-off knob in this practical range.}
    \label{tab:tau-comp-sweep}
\end{table*}

\section{Additional Table-Selection Analyses}
\label{app:table-selection-analysis}


\paragraph{Sensitivity of the adjustment threshold $\tau_{\text{comp}}$.}
In \textsc{CORE-T}, the additive adjustment step re-introduces candidate tables whose cached compatibility with the selected set exceeds a threshold $\tau_{\text{comp}}$.
Appendix Table~\ref{tab:tau-comp-sweep} sweeps $\tau_{\text{comp}}\in\{0.3,0.5,0.7\}$ across \textsc{Bird}, \textsc{Spider}, \textsc{MMQA}, and \textsc{Beaver} for both selector LLMs.
Across settings, increasing $\tau_{\text{comp}}$ generally returns fewer tables while slightly improving F1 and slightly reducing perfect recall (PR), consistent with a precision--recall/compactness trade-off. The changes are modest in this range: F1 varies by at most 4.1 percentage points, PR by at most 1.4 percentage points, and the average number of returned tables by at most 0.4.
Thus, $\tau_{\text{comp}}$ behaves as an interpretable trade-off knob rather than a brittle heuristic.

\paragraph{Prompt length analysis (selector input tokens).}
To assess whether long prompts harm the selector's reasoning, we measure the selector input token budget under the fixed 5-row cap and analyze performance as a function of prompt length.
Appendix Table~\ref{tab:promptlen-stats} reports min/median/max tokens across datasets and selector LLMs; \textsc{Beaver} produces the longest prompts (up to 27{,}894 tokens with Qwen-2.5-7B), while \textsc{Spider} and \textsc{MMQA} remain substantially shorter.
We further bin queries into prompt-length tertiles (short/medium/long by selector input tokens) and report full-pipeline table-selection metrics (F1 and PR) per bin in Appendix Table~\ref{tab:promptlen-bins}.
The tertile results show some length sensitivity, especially on \textsc{Beaver}, where the long-vs.-short shift reaches $-22.4$ F1 points with Qwen-2.5-7B and $-22.9$ PR points with Llama-3.1-8B. Other datasets show smaller or non-monotonic changes (e.g., \textsc{Bird}/Qwen improves from 70.0 to 72.4 F1 from short to long), suggesting that prompt length alone does not fully explain selection quality.

\begin{table*}[!t]
    \centering
    \small
    \setlength{\tabcolsep}{4pt}
    \renewcommand{\arraystretch}{1.15}
    \begin{tabular}{l l
      c c
      c c
      c c}
    \toprule
    \textbf{Selector} & \textbf{Dataset} &
    \multicolumn{2}{c}{\textbf{Short tertile}} &
    \multicolumn{2}{c}{\textbf{Medium tertile}} &
    \multicolumn{2}{c}{\textbf{Long tertile}} \\
    \cmidrule(lr){3-4}\cmidrule(lr){5-6}\cmidrule(lr){7-8}
    & &
    \textbf{F1 (\%)} & \textbf{PR (\%)} &
    \textbf{F1 (\%)} & \textbf{PR (\%)} &
    \textbf{F1 (\%)} & \textbf{PR (\%)} \\
    \midrule
    
    \multirow{4}{*}{\textbf{Llama-3.1-8B}} &
    \textsc{Bird}   & 64.1 & 88.7 & 63.3 & 90.2 & 59.5 & 91.0 \\
    & \textsc{Spider} & 58.0 & 95.4 & 53.9 & 97.4 & 49.4 & 97.1 \\
    & \textsc{MMQA}   & 51.4 & 61.0 & 50.4 & 66.3 & 45.6 & 56.5 \\
    & \textsc{Beaver} & 49.4 & 28.6 & 36.8 & 11.6 & 32.3 & 5.7 \\
    \midrule
    
    \multirow{4}{*}{\textbf{Qwen-2.5-7B}} &
    \textsc{Bird}   & 70.0 & 86.5 & 74.7 & 87.6 & 72.4 & 86.7 \\
    & \textsc{Spider} & 68.5 & 96.8 & 64.7 & 94.2 & 67.1 & 93.6 \\
    & \textsc{MMQA}   & 60.2 & 62.3 & 58.3 & 63.3 & 50.7 & 52.4 \\
    & \textsc{Beaver} & 53.7 & 25.7 & 44.0 & 18.8 & 31.3 & 8.6 \\
    \bottomrule
    \end{tabular}
    
    \caption{\textbf{Prompt-length tertiles vs.\ full-pipeline table-selection performance.}
    Queries are binned into short/medium/long tertiles by selector input tokens (computed under the fixed 5-row snapshot). We report the resulting full-pipeline F1 and perfect recall (PR) after adjustment ($\tau_{\text{comp}}{=}0.5$).}
    \label{tab:promptlen-bins}
\end{table*}


\paragraph{Gold-table drop and recovery by adjustment.}
To quantify how often the selector discards required tables and how often adjustment recovers them, we report query-level gold-table drop and recovery rates in Appendix Table~\ref{tab:gold-recovery}.
Across datasets and both selector LLMs, the selector drops at least one gold table in 4.7--42.1\% of queries, with the highest drop rates on \textsc{Beaver}. Conditional on such dropped cases, adjustment recovers at least one missing gold table in 9.1--57.1\% of cases and recovers all dropped gold tables in 9.1--50.8\% of cases, indicating that the additive step often repairs selection mistakes but is less effective on the hardest \textsc{Beaver} setting.

\begin{table*}[!t]
    \centering
    \small
    \setlength{\tabcolsep}{3pt}
    \renewcommand{\arraystretch}{1.1}
    \begin{tabular}{l l
      c c c
      c c c
      c c c}
    \toprule
    \textbf{Selector} & \textbf{Dataset} &
    \multicolumn{3}{c}{\textbf{Sel.\ drops $\ge 1$}} &
    \multicolumn{3}{c}{\textbf{Adj.\ rec.\ $\ge 1$}} &
    \multicolumn{3}{c}{\textbf{Adj.\ rec.\ ALL}} \\
    \cmidrule(lr){3-5}\cmidrule(lr){6-8}\cmidrule(lr){9-11}
    & &
    \textbf{\%} & \textbf{Count} & \textbf{Total} &
    \textbf{\%} & \textbf{Count} & \textbf{Total} &
    \textbf{\%} & \textbf{Count} & \textbf{Total} \\
    \midrule
    
    \multirow{4}{*}{\textbf{Llama-3.1-8B}} &
    \textsc{Bird}   & 10.4\% & 160 & 1534 & 50.0\% & 80  & 160 & 48.1\% & 77  & 160 \\
    & \textsc{Spider} & 4.7\%  & 49  & 1034 & 46.9\% & 23  & 49  & 46.9\% & 23  & 49  \\
    & \textsc{MMQA}   & 12.9\% & 142 & 1105 & 31.0\% & 44  & 142 & 30.3\% & 43  & 142 \\
    & \textsc{Beaver} & 42.1\% & 88  & 209  & 9.1\%  & 8   & 88  & 9.1\%  & 8   & 88  \\
    \midrule
    
    \multirow{4}{*}{\textbf{Qwen-2.5-7B}} &
    \textsc{Bird}   & 16.6\% & 254 & 1534 & 57.1\% & 145 & 254 & 50.8\% & 129 & 254 \\
    & \textsc{Spider} & 8.6\%  & 89  & 1034 & 48.3\% & 43  & 89  & 47.2\% & 42  & 89  \\
    & \textsc{MMQA}   & 20.5\% & 227 & 1105 & 49.8\% & 113 & 227 & 47.1\% & 107 & 227 \\
    & \textsc{Beaver} & 39.2\% & 82  & 209  & 26.8\% & 22  & 82  & 14.6\% & 12  & 82  \\
    \bottomrule
    \end{tabular}
    
    \caption{\textbf{Gold-table drop and recovery (query level, $\tau_{\text{comp}}{=}0.5$).}
    We report (i) the fraction of queries where the selector drops $\ge 1$ gold table, and (ii) conditional on dropped cases, the fraction where adjustment recovers $\ge 1$ or all dropped gold tables.
    \textbf{Count/Total} gives the raw numerator/denominator for each percentage.}
    \label{tab:gold-recovery}
\end{table*}


\paragraph{Robustness to semantically weak join-critical tables ("bridge tables").}
A key challenge in pooled open-book corpora is that some gold tables are \emph{join-critical} but have low surface semantic relevance to the query (e.g., connector/bridge tables in a join path).
To stress-test this failure mode, we analyze \emph{bridge tables} defined as gold tables whose relevance score to the query is $\le 0.5$ (on \textsc{Bird}, $0.5$ lies near the low-score tail of gold-table relevance (q20=0.493, q25=0.507)).
Across \textsc{Bird}/\textsc{Spider}/\textsc{MMQA} and both selector LLMs, the selector drops at least one bridge table in 9.2--27.3\% of applicable bridge queries. Conditional on such drops, the adjustment step recovers $\ge 1$ dropped bridge table in 22.5--57.1\% of affected queries and fully recovers all dropped bridge tables in 22.5--57.1\%.
Appendix Table~\ref{tab:bridge-recovery} reports the full drop and recovery rates (with counts) for $\tau_{\text{comp}}{=}0.5$; \textsc{Beaver} has no applicable dropped bridge cases under this definition.

\begin{table*}[!t]
    \centering
    \small
    \setlength{\tabcolsep}{3pt}
    \renewcommand{\arraystretch}{1.1}
    \begin{tabular}{l l
        c c c
        c c c
        c c c}
    \toprule
    \textbf{Selector} & \textbf{Dataset} &
    \multicolumn{3}{c}{\textbf{Sel.\ drops $\ge 1$}} &
    \multicolumn{3}{c}{\textbf{Adj.\ rec.\ $\ge 1$}} &
    \multicolumn{3}{c}{\textbf{Adj.\ rec.\ ALL}} \\
    \cmidrule(lr){3-5}\cmidrule(lr){6-8}\cmidrule(lr){9-11}
    & &
    \textbf{\%} & \textbf{Count} & \textbf{Total} &
    \textbf{\%} & \textbf{Count} & \textbf{Total} &
    \textbf{\%} & \textbf{Count} & \textbf{Total} \\
    \midrule
    
    \multirow{4}{*}{\textbf{Llama-3.1-8B}} &
    \textsc{Bird}   & 20.4\% & 91  & 445 & 41.8\% & 38 & 91 & 39.6\% & 36 & 91 \\
    & \textsc{Spider} & 9.2\%  & 14  & 152 & 57.1\% & 8  & 14 & 57.1\% & 8  & 14 \\
    & \textsc{MMQA}   & 16.1\% & 40  & 248 & 22.5\% & 9  & 40 & 22.5\% & 9  & 40 \\
    & \textsc{Beaver} & 0.0\%  & 0   & 0   & 0.0\%  & 0  & 0  & 0.0\%  & 0  & 0  \\
    \midrule
    
    \multirow{4}{*}{\textbf{Qwen-2.5-7B}} &
    \textsc{Bird}   & 22.2\% & 90  & 406 & 54.4\% & 49 & 90 & 47.8\% & 43 & 90 \\
    & \textsc{Spider} & 27.3\% & 35  & 128 & 45.7\% & 16 & 35 & 45.7\% & 16 & 35 \\
    & \textsc{MMQA}   & 26.9\% & 58  & 216 & 48.3\% & 28 & 58 & 48.3\% & 28 & 58 \\
    & \textsc{Beaver} & 0.0\%  & 0   & 2   & 0.0\%  & 0  & 0  & 0.0\%  & 0  & 0  \\
    \bottomrule
    \end{tabular}
    \caption{\textbf{Bridge-table drop and recovery (query level, $\tau_{\text{comp}}{=}0.5$).}
    We report (i) the fraction of bridge queries where the selector drops $\ge 1$ bridge gold table, and (ii) conditional on dropped cases, the fraction where adjustment recovers $\ge 1$ or all dropped bridge tables.
    \textbf{Count/Total} gives the raw numerator/denominator for each percentage; $0.0\%$ with total $0$ indicates no applicable bridge queries or no dropped bridge cases.
    Denominators for “Sel.\ drops $\ge 1$” are queries that contain at least one bridge gold table and whose DR@10 (initial table retrieval step) candidate set includes all bridge gold tables (so the selector can drop them).}
    \label{tab:bridge-recovery}
\end{table*}

\begin{table*}[!t]
    \centering
    \small
    \setlength{\tabcolsep}{3pt}            
    \renewcommand{\arraystretch}{1.1}      
    \begin{adjustbox}{max width=\textwidth}
    \begin{tabular}{
        @{} L{1.75cm}
        C{1.15cm} C{0.50cm} C{0.52cm} C{0.52cm}
        >{\centering\arraybackslash}m{4pt}
        C{1.15cm} C{0.50cm} C{0.52cm} C{0.52cm}
        >{\centering\arraybackslash}m{4pt}
        C{1.15cm} C{0.50cm} C{0.52cm} C{0.52cm}
        >{\centering\arraybackslash}m{4pt}
        C{1.15cm} C{0.50cm} C{0.52cm} C{0.52cm}
        @{}
    }
    & \multicolumn{4}{c}{\textbf{Bird} (n=1534, $\bar G$=1.95)} && \multicolumn{4}{c}{\textbf{Spider} (n=1034, $\bar G$=1.51)} && \multicolumn{4}{c}{\textbf{MMQA} (n=1105, $\bar G$=2.20)} && \multicolumn{4}{c}{\textbf{Beaver} (n=209, $\bar G$=4.44)} \\
    \cmidrule(lr){2-5} \cmidrule(lr){7-10} \cmidrule(lr){12-15} \cmidrule(lr){17-20}
    \textbf{} & {\scriptsize Avg.\#tab.$\downarrow$} & \textbf{P} & \textbf{F1} & \textbf{PR}
    & & {\scriptsize Avg.\#tab.$\downarrow$} & \textbf{P} & \textbf{F1} & \textbf{PR}
    & & {\scriptsize Avg.\#tab.$\downarrow$} & \textbf{P} & \textbf{F1} & \textbf{PR}
    & & {\scriptsize Avg.\#tab.$\downarrow$} & \textbf{P} & \textbf{F1} & \textbf{PR} \\
    \midrule
    \multicolumn{20}{l}{\textbf{UAE-Large-V1}} \\
    \midrule
    DR@5 & 5.0 & 34.9 & 49.4 & 83.0 && 5.0 & 29.4 & 43.9 & 95.6  && 5.0 & 30.8 & 42.6 & 50.3  && 5.0 & 33.9 & 36.5 & 13.9 \\
    DR@8 & 8.0 & 23.1 & 36.5 & 91.5 && 8.0 & 18.8 & 30.9 & 98.9  && 8.0 & 21.2 & 33.1 & 61.3  && 8.0 & 26.9 & 34.2 & 20.6 \\
    \rowdash
    DR@10 & 10.0 & 18.8 & 31.0 & 94.3 && 10.0 & 15.0 & 25.7 & 99.1  && 10.0 & 17.6 & 28.7 & 66.2  && 10.0 & 24.0 & 32.7 & 26.3 \\
    \rowdash
    DR@15 & 15.0 & 12.7 & 22.3 & 97.0 && 15.0 & 10.1 & 18.0 & 99.7  && 15.0 & 12.4 & 21.5 & 73.1  && 15.0 & 18.8 & 28.5 & 32.5 \\
    \addlinespace[4pt]
    \midrule
    \multicolumn{20}{l}{\textbf{Snowflake-arctic-embed-m-v2.0}} \\
    \midrule
    DR@5 & 5.0 & 33.2 & 47.1 & 76.2 && 5.0 & 29.1 & 43.5 & 94.3  && 5.0 & 31.3 & 43.3 & 49.0  && 5.0 & 36.3 & 39.2 & 12.4 \\
    DR@8 & 8.0 & 22.2 & 35.2 & 85.7 && 8.0 & 18.6 & 30.5 & 97.1  && 8.0 & 21.3 & 33.3 & 59.9  && 8.0 & 29.0 & 37.4 & 24.9 \\
    DR@10 & 10.0 & 18.2 & 30.0 & 88.8 && 10.0 & 14.9 & 25.4 & 97.9  && 10.0 & 17.5 & 28.5 & 63.3  && 10.0 & 25.5 & 35.2 & 30.6 \\
    DR@15 & 15.0 & 12.4 & 21.7 & 92.6 && 15.0 & 10.0 & 17.9 & 98.4  && 15.0 & 12.2 & 21.1 & 69.4  && 15.0 & 19.7 & 29.9 & 36.8 \\
    \addlinespace[4pt]
    \midrule
    \multicolumn{20}{l}{\textbf{text-embedding-3-large}} \\
    \midrule
    DR@5 & 5.0 & 35.9 & 50.7 & 86.9 && 5.0 & 29.7 & 44.4 & 97.4  && 5.0 & 34.0 & 47.0 & 61.8  && 5.0 & 39.3 & 42.5 & 16.7 \\
    DR@8 & 8.0 & 23.4 & 37.1 & 93.7 && 8.0 & 18.8 & 30.9 & 98.8  && 8.0 & 23.0 & 35.9 & 72.4  && 8.0 & 31.3 & 40.1 & 28.2 \\
    DR@10 & 10.0 & 19.0 & 31.4 & 96.3 && 10.0 & 15.1 & 25.7 & 99.3  && 10.0 & 18.9 & 30.9 & 76.8  && 10.0 & 27.3 & 37.3 & 32.5 \\
    DR@15 & 15.0 & 12.8 & 22.5 & 98.2 && 15.0 & 10.1 & 18.1 & 99.9  && 15.0 & 13.1 & 22.8 & 82.5  && 15.0 & 21.3 & 32.2 & 40.2 \\
    \addlinespace[4pt]
    \bottomrule
    \end{tabular}%
    \end{adjustbox}
    \caption{\textbf{Initial table retrieval performance at different top-$K$ cutoffs.}
    Each block is an embedding model. We report average retrieved tables, precision (P), F1, and perfect recall (PR), where PR is the percentage of questions for which all gold tables are retrieved. \(\bar G\) is the average number of gold tables per query. \emph{The row between the dashed rules indicates the configuration we adopt for this step and carry forward to the subsequent step.}
    All methods used Llama3.1-8B-Instruct as the metadata (table purpose) generator.}
    \label{tab:dr-topk-results}
\end{table*}

\begin{table*}[!t]
    \centering
    \small
    \setlength{\tabcolsep}{2pt}
    \renewcommand{\arraystretch}{1.08}
    \begin{adjustbox}{max width=\textwidth}
    \begin{tabular}{
      @{} L{2.05cm}
      c c c
      @{\hspace{5pt}}
      c c c
      @{\hspace{5pt}}
      c c c
      @{\hspace{5pt}}
      c c c
      @{}
    }
    & \multicolumn{3}{c}{\textbf{Bird} (n=1534)} & \multicolumn{3}{c}{\textbf{Spider} (n=1034)} & \multicolumn{3}{c}{\textbf{MMQA} (n=1105)} & \multicolumn{3}{c}{\textbf{Beaver} (n=209)} \\
    \cmidrule(lr){2-4}
    \cmidrule(lr){5-7}
    \cmidrule(lr){8-10}
    \cmidrule(l){11-13}
     & \makecell[c]{EX$_{MT}$\\(76.4\%)} & \makecell[c]{EX$_{all}$\\(100\%)} & \makecell[c]{EX$_{PR}$\\(87.0\%)} & \makecell[c]{EX$_{MT}$\\(44.4\%)} & \makecell[c]{EX$_{all}$\\(100\%)} & \makecell[c]{EX$_{PR}$\\(94.9\%)} & \makecell[c]{EX$_{MT}$\\(99.6\%)} & \makecell[c]{EX$_{all}$\\(100\%)} & \makecell[c]{EX$_{PR}$\\(59.4\%)} & \makecell[c]{EX$_{MT}$\\(98.6\%)} & \makecell[c]{EX$_{all}$\\(100\%)} & \makecell[c]{EX$_{PR}$\\(17.7\%)} \\
    \midrule
    \multicolumn{13}{l}{\textbf{Llama-3.2-3B}} \\
    \midrule
    DR@5 & 2.1 & 3.1 & 2.9 & 9.6 & 14.8 & 14.8 & 2.1 & 2.1 & 1.7 & 0.0 & 0.0 & \underline{0.0} \\
    ReAct & 8.6 & \underline{12.6} & \underline{12.5} & 19.8 & 34.1 & 34.0 & 5.7 & 5.7 & 5.2 & 0.0 & 0.0 & \underline{0.0} \\
    JAR@5 & 3.3 & 4.2 & 4.2 & 11.1 & 14.5 & 14.5 & — & — & — & — & — & — \\
    REAR & \underline{9.6} & 10.0 & 9.8 & \underline{26.4} & \underline{29.6} & \underline{29.3} & \underline{8.4} & \underline{8.3} & \underline{6.9} & \underline{1.0} & \underline{1.0} & \underline{0.0} \\
    ARM & 7.3 & 8.7 & 8.5 & — & — & — & — & — & — & — & — & — \\
    \rowdash
    CORE-T & \textbf{17.2} & \textbf{18.5} & \textbf{18.1} & \textbf{37.5} & \textbf{41.0} & \textbf{40.7} & \textbf{19.3} & \textbf{19.2} & \textbf{17.1} & \textbf{2.4} & \textbf{2.4} & \textbf{1.4} \\
    \rowdash
    Oracle & 24.8 & 27.8 & 27.8 & 45.1 & 58.9 & 58.9 & 45.4 & 45.4 & 45.4 & 1.5 & 1.9 & 1.9 \\
    \addlinespace[4pt]
    \midrule
    \multicolumn{13}{l}{\textbf{Gemma-3-4B}} \\
    \midrule
    DR@5 & 14.9 & 19.8 & 18.7 & 39.9 & 52.4 & 52.3 & 19.3 & 19.3 & 15.7 & \underline{3.9} & \underline{3.8} & \textbf{1.0} \\
    ReAct & 14.5 & 20.7 & 19.9 & \underline{40.7} & \textbf{54.3} & \textbf{54.0} & \underline{22.8} & \underline{22.8} & \underline{19.5} & \underline{3.9} & \underline{3.8} & \textbf{1.0} \\
    JAR@5 & 17.5 & 22.1 & 21.3 & 40.5 & 51.3 & 51.2 & — & — & — & — & — & — \\
    REAR & 15.9 & 20.1 & 19.6 & 35.3 & 45.1 & 44.7 & 16.1 & 16.1 & 12.9 & \textbf{4.9} & \textbf{4.8} & \underline{0.5} \\
    ARM & \underline{17.7} & \underline{22.6} & \underline{21.6} & — & — & — & — & — & — & — & — & — \\
    \rowdash
    CORE-T & \textbf{19.5} & \textbf{24.6} & \textbf{23.6} & \textbf{44.4} & \underline{53.6} & \underline{53.2} & \textbf{25.2} & \textbf{25.2} & \textbf{23.1} & 2.4 & 2.9 & \textbf{1.0} \\
    \rowdash
    Oracle & 24.7 & 30.4 & 30.4 & 53.8 & 65.7 & 65.7 & 47.6 & 47.5 & 47.5 & 4.4 & 4.8 & 4.8 \\
    \addlinespace[4pt]
    \midrule
    \multicolumn{13}{l}{\textbf{GPT-4o-mini}} \\
    \midrule
    DR@5 & 34.9 & 40.3 & 38.3 & 53.4 & 65.0 & 64.7 & 32.2 & 32.3 & 28.9 & \textbf{7.8} & \textbf{8.1} & \underline{2.4} \\
    ReAct & 35.0 & 40.4 & 38.7 & 53.6 & \underline{65.7} & \underline{65.3} & \underline{34.5} & \underline{34.6} & \underline{30.6} & \underline{6.3} & \underline{6.7} & \textbf{2.9} \\
    JAR@5 & 36.3 & 41.3 & 39.9 & \underline{55.1} & \textbf{65.8} & \textbf{65.5} & — & — & — & — & — & — \\
    REAR & \underline{38.4} & 42.3 & \underline{40.6} & 52.3 & 62.3 & 61.7 & 28.9 & 29.0 & 25.2 & 4.4 & 4.3 & 0.5 \\
    ARM & 37.8 & \underline{42.4} & \underline{40.6} & — & — & — & — & — & — & — & — & — \\
    \rowdash
    CORE-T & \textbf{38.6} & \textbf{43.2} & \textbf{41.7} & \textbf{55.3} & \textbf{65.8} & \underline{65.3} & \textbf{36.1} & \textbf{36.0} & \textbf{33.1} & 5.8 & 6.2 & \underline{2.4} \\
    \rowdash
    Oracle & 47.8 & 50.7 & 50.7 & 64.5 & 71.8 & 71.8 & 65.8 & 65.7 & 65.7 & 6.8 & 7.2 & 7.2 \\
    \addlinespace[4pt]
    \addlinespace[0.5pt]
    \bottomrule
    \end{tabular}
    \end{adjustbox}
    \caption{\textbf{End-to-end SQL execution performance with Qwen-2.5-7B-Instruct as the table selector.}
    Execution accuracy (EX) on \textsc{Bird}, \textsc{Spider}, \textsc{MMQA}, and \textsc{Beaver} for multi-table queries (EX$_{MT}$, $\ge 2$ gold tables), all queries (EX$_{all}$), and the perfect-recall subset (EX$_{PR}$), where all gold tables required by the gold SQL are retrieved and included among the tables passed to the SQL generator. Oracle uses gold tables. \textbf{Best}/\underline{second best} are among non-oracle methods within each SQL-generator block.}
    \label{tab:answer-gen-table-qwen}
\end{table*}

\section{Error Analysis Details}
\label{app:error-analysis}

\begin{table}[!t]
    \centering
    \small
    \setlength{\tabcolsep}{5pt}
    \renewcommand{\arraystretch}{1.1}
    \begin{adjustbox}{max width=\columnwidth}
    \begin{tabular}{@{}lcccc@{}}
    & \multicolumn{2}{c}{\textbf{Table Selection}} & \multicolumn{2}{c}{\textbf{SQL Generation}} \\
    \cmidrule(lr){2-3}\cmidrule(lr){4-5}
    & \makecell{\textbf{Recall}\\\textbf{issue}$\downarrow$} &
      \makecell{\textbf{Precision}\\\textbf{issue}$\downarrow$} &
      \makecell{\textbf{Schema}\\\textbf{linking}$\downarrow$} &
      \makecell{\textbf{Formatting}$\downarrow$} \\
    \midrule
    \textsc{ARM} &
    \textbf{291} (19.0\%) &
    1112 (72.5\%) &
    \textbf{37} (2.4\%) &
    114 (7.4\%) \\
    \rowdash
    \textsc{CORE-T} &
    347 (22.6\%) &
    \textbf{855} (55.7\%) &
    107 (7.0\%) &
    \textbf{2} (0.1\%) \\
    \bottomrule
    \end{tabular}
    \end{adjustbox}
    \caption{\textbf{Error breakdown on \textsc{Bird} (n=1534).}
    Results use Llama-3.1-8B-Instruct as table selector and Llama-3.2-3B-Instruct as SQL generator.
    We report multi-label error counts and rates over all queries for table selection errors computed from gold vs.\ selected table sets. SQL-only categories (schema linking and formatting) are assigned only when the generated SQL uses all and only the gold tables. Categories are multi-label (a query may contribute to multiple categories). Bold indicates the lower value.}
    \label{tab:error-analysis-bird}
\end{table}

\paragraph{Paired significance tests.}
Table-selection F1 is continuous and paired across the same questions, so we use a two-sided paired sign-flip test with 10{,}000 randomizations, avoiding a parametric normality assumption. For binary paired EX, we use the exact McNemar test, which compares queries solved only by \textsc{CORE-T} with those solved only by the baseline. We consider $p<.05$ significant. Across all available settings, \textsc{CORE-T} is significantly better in 17 of 19 F1 comparisons; one comparison is a tie, and only \textsc{Spider} against \textsc{ReAct} significantly favors the baseline. Across 57 EX comparisons over four datasets and three SQL generators, \textsc{CORE-T} is significantly better in 30 and tied in 27, and is never significantly worse. Tables~\ref{tab:paired-f1-all} and~\ref{tab:paired-ex-all} report every available paired comparison.

\begin{table}[!t]
    \centering
    \small
    \setlength{\tabcolsep}{3pt}
    \renewcommand{\arraystretch}{0.98}
    \begin{adjustbox}{max width=\columnwidth}
    \begin{tabular}{@{}lrrl@{}}
    \toprule
     & \textbf{$\Delta$ F1} & \textbf{$p$} & \textbf{Winner} \\
    \midrule
    \multicolumn{4}{l}{\textbf{\textsc{Bird}}} \\
    \midrule
    ARM & +4.1 & $<10^{-4}$ & \textsc{CORE-T} \\
    DR@5 & +22.6 & $<10^{-4}$ & \textsc{CORE-T} \\
    DR@10 & +41.4 & $<10^{-4}$ & \textsc{CORE-T} \\
    JAR@5 & +22.2 & $<10^{-4}$ & \textsc{CORE-T} \\
    ReAct & +0.7 & $.292$ & Tie \\
    REAR & +21.4 & $<10^{-4}$ & \textsc{CORE-T} \\
    \addlinespace[2pt]
    \midrule
    \multicolumn{4}{l}{\textbf{\textsc{Spider}}} \\
    \midrule
    DR@5 & +22.6 & $<10^{-4}$ & \textsc{CORE-T} \\
    DR@10 & +41.1 & $<10^{-4}$ & \textsc{CORE-T} \\
    JAR@5 & +22.7 & $<10^{-4}$ & \textsc{CORE-T} \\
    ReAct & $-8.7$ & $<10^{-4}$ & ReAct \\
    REAR & +24.7 & $<10^{-4}$ & \textsc{CORE-T} \\
    \addlinespace[2pt]
    \midrule
    \multicolumn{4}{l}{\textbf{\textsc{MMQA}}} \\
    \midrule
    DR@5 & +13.4 & $<10^{-4}$ & \textsc{CORE-T} \\
    DR@10 & +27.5 & $<10^{-4}$ & \textsc{CORE-T} \\
    ReAct & +6.0 & $<10^{-4}$ & \textsc{CORE-T} \\
    REAR & +17.2 & $<10^{-4}$ & \textsc{CORE-T} \\
    \addlinespace[2pt]
    \midrule
    \multicolumn{4}{l}{\textbf{\textsc{Beaver}}} \\
    \midrule
    DR@5 & +5.7 & $<10^{-4}$ & \textsc{CORE-T} \\
    DR@10 & +9.6 & $<10^{-4}$ & \textsc{CORE-T} \\
    ReAct & +17.7 & $<10^{-4}$ & \textsc{CORE-T} \\
    REAR & +7.5 & $<10^{-4}$ & \textsc{CORE-T} \\
    \bottomrule
    \end{tabular}
    \end{adjustbox}
    \caption{\textbf{Paired F1 significance across all 19 available dataset--baseline comparisons.}
    $\Delta$ is the F1 difference in percentage points (\textsc{CORE-T}$-$baseline). The winner is determined at $p<.05$.}
    \label{tab:paired-f1-all}
\end{table}

\paragraph{BIRD-stratified analysis.}
\textsc{Bird} is the only evaluated dataset with per-query difficulty labels. We test seven strata: difficulty (simple, moderate, challenging) and gold-table count (1, 2, 3, 4+), applying Holm correction separately across the seven F1 and seven EX tests. Tables~\ref{tab:bird-stratified-f1} and~\ref{tab:bird-stratified-ex} report the number of strata whose Holm-adjusted $p$-value remains below .05. The Winner column instead summarizes the overall paired comparison. With Llama-3.2-3B, all seven F1 strata and six of seven EX strata remain significant against DR@5 and JAR@5, showing that the gains are not confined to easy or low-table-count questions.

\begin{table}[!t]
    \centering
    \small
    \setlength{\tabcolsep}{3pt}
    \renewcommand{\arraystretch}{1.0}
    \begin{adjustbox}{max width=\columnwidth}
    \begin{tabular}{@{}lrrrl@{}}
    \toprule
     & \textbf{$\Delta$ F1} & \textbf{$p$} & \textbf{Strata} & \textbf{Winner} \\
    \midrule
    \multicolumn{5}{l}{\textbf{\textsc{Bird}}} \\
    \midrule
    ARM & +4.1 & $<10^{-4}$ & 3/7 & \textsc{CORE-T} \\
    DR@5 & +22.6 & $<10^{-4}$ & 7/7 & \textsc{CORE-T} \\
    DR@10 & +41.4 & $<10^{-4}$ & 7/7 & \textsc{CORE-T} \\
    JAR@5 & +22.2 & $<10^{-4}$ & 7/7 & \textsc{CORE-T} \\
    ReAct & +0.7 & $.292$ & 7/7 & Tie \\
    REAR & +21.4 & $<10^{-4}$ & 7/7 & \textsc{CORE-T} \\
    \bottomrule
    \end{tabular}
    \end{adjustbox}
    \caption{\textbf{Paired and BIRD-stratified F1 significance.}
    $\Delta$ is the F1 difference in percentage points. Strata reports how many of the seven subgroup tests have Holm-adjusted $p<.05$; it records significance, not effect direction. Winner refers to the overall paired test.}
    \label{tab:bird-stratified-f1}
\end{table}

\begin{table}[!t]
    \centering
    \small
    \setlength{\tabcolsep}{3pt}
    \renewcommand{\arraystretch}{0.98}
    \begin{adjustbox}{max width=\columnwidth}
    \begin{tabular}{@{}lrrrl@{}}
    \toprule
     & \textbf{$\Delta$ EX} & \textbf{$p$} & \textbf{Strata} & \textbf{Winner} \\
    \midrule
    \multicolumn{5}{l}{\textbf{Llama-3.2-3B}} \\
    \midrule
    ARM & +9.8 & $<10^{-3}$ & 6/7 & \textsc{CORE-T} \\
    DR@5 & +15.4 & $<10^{-3}$ & 6/7 & \textsc{CORE-T} \\
    DR@10 & +5.7 & $<10^{-3}$ & 4/7 & \textsc{CORE-T} \\
    JAR@5 & +14.3 & $<10^{-3}$ & 6/7 & \textsc{CORE-T} \\
    ReAct & +5.9 & $<10^{-3}$ & 5/7 & \textsc{CORE-T} \\
    REAR & +8.5 & $<10^{-3}$ & 4/7 & \textsc{CORE-T} \\
    \addlinespace[2pt]
    \midrule
    \multicolumn{5}{l}{\textbf{GPT-4o-mini}} \\
    \midrule
    ARM & +0.7 & $.507$ & 0/7 & Tie \\
    DR@5 & +2.9 & $2\!\times\!10^{-3}$ & 2/7 & \textsc{CORE-T} \\
    DR@10 & +0.6 & $.543$ & 0/7 & Tie \\
    JAR@5 & +1.8 & $.063$ & 0/7 & Tie \\
    ReAct & +2.7 & $3\!\times\!10^{-3}$ & 1/7 & \textsc{CORE-T} \\
    REAR & +0.8 & $.438$ & 0/7 & Tie \\
    \addlinespace[2pt]
    \midrule
    \multicolumn{5}{l}{\textbf{Gemma-3-4B}} \\
    \midrule
    ARM & +2.0 & $.048$ & 0/7 & \textsc{CORE-T} \\
    DR@5 & +4.8 & $<10^{-3}$ & 2/7 & \textsc{CORE-T} \\
    DR@10 & +6.1 & $<10^{-3}$ & 3/7 & \textsc{CORE-T} \\
    JAR@5 & +2.5 & $2\!\times\!10^{-2}$ & 0/7 & \textsc{CORE-T} \\
    ReAct & +3.8 & $<10^{-3}$ & 2/7 & \textsc{CORE-T} \\
    REAR & +4.5 & $<10^{-3}$ & 3/7 & \textsc{CORE-T} \\
    \bottomrule
    \end{tabular}
    \end{adjustbox}
    \caption{\textbf{Paired and BIRD-stratified EX$_{all}$ significance.}
    $\Delta$ is the EX$_{all}$ difference in percentage points. Strata reports how many of the seven subgroup tests have Holm-adjusted $p<.05$; it records significance, not effect direction. Winner refers to the overall paired test.}
    \label{tab:bird-stratified-ex}
\end{table}

\begin{table*}[!t]
    \centering
    \small
    \setlength{\tabcolsep}{2.5pt}
    \renewcommand{\arraystretch}{0.98}
    \begin{adjustbox}{max width=\textwidth}
    \begin{tabular}{@{}lrrlrrlrrl@{}}
    & \multicolumn{3}{c}{\textbf{Llama-3.2-3B}} & \multicolumn{3}{c}{\textbf{GPT-4o-mini}} & \multicolumn{3}{c}{\textbf{Gemma-3-4B}} \\
    \cmidrule(lr){2-4}\cmidrule(lr){5-7}\cmidrule(lr){8-10}
     & \textbf{$\Delta$} & \textbf{$p$} & \textbf{Winner} & \textbf{$\Delta$} & \textbf{$p$} & \textbf{Winner} & \textbf{$\Delta$} & \textbf{$p$} & \textbf{Winner} \\
    \midrule
    \multicolumn{10}{l}{\textbf{\textsc{Bird}}} \\
    \midrule
    ARM & +9.8 & $<10^{-3}$ & \textsc{CORE-T} & +0.7 & $.507$ & Tie & +2.0 & $.048$ & \textsc{CORE-T} \\
    DR@5 & +15.4 & $<10^{-3}$ & \textsc{CORE-T} & +2.9 & $2\!\times\!10^{-3}$ & \textsc{CORE-T} & +4.8 & $<10^{-3}$ & \textsc{CORE-T} \\
    DR@10 & +5.7 & $<10^{-3}$ & \textsc{CORE-T} & +0.6 & $.543$ & Tie & +6.1 & $<10^{-3}$ & \textsc{CORE-T} \\
    JAR@5 & +14.3 & $<10^{-3}$ & \textsc{CORE-T} & +1.8 & $.063$ & Tie & +2.5 & $2\!\times\!10^{-2}$ & \textsc{CORE-T} \\
    ReAct & +5.9 & $<10^{-3}$ & \textsc{CORE-T} & +2.7 & $3\!\times\!10^{-3}$ & \textsc{CORE-T} & +3.8 & $<10^{-3}$ & \textsc{CORE-T} \\
    REAR & +8.5 & $<10^{-3}$ & \textsc{CORE-T} & +0.8 & $.438$ & Tie & +4.5 & $<10^{-3}$ & \textsc{CORE-T} \\
    \addlinespace[2pt]
    \midrule
    \multicolumn{10}{l}{\textbf{\textsc{Spider}}} \\
    \midrule
    DR@5 & +26.2 & $<10^{-3}$ & \textsc{CORE-T} & +0.8 & $.451$ & Tie & +1.2 & $.369$ & Tie \\
    DR@10 & +5.5 & $<10^{-3}$ & \textsc{CORE-T} & $-1.2$ & $.219$ & Tie & +2.4 & $.090$ & Tie \\
    JAR@5 & +26.5 & $<10^{-3}$ & \textsc{CORE-T} & +0.0 & $1.000$ & Tie & +2.3 & $.086$ & Tie \\
    ReAct & +6.9 & $<10^{-3}$ & \textsc{CORE-T} & +0.1 & $1.000$ & Tie & $-0.7$ & $.628$ & Tie \\
    REAR & +11.4 & $<10^{-3}$ & \textsc{CORE-T} & +3.5 & $2\!\times\!10^{-3}$ & \textsc{CORE-T} & +8.5 & $<10^{-3}$ & \textsc{CORE-T} \\
    \addlinespace[2pt]
    \midrule
    \multicolumn{10}{l}{\textbf{\textsc{MMQA}}} \\
    \midrule
    DR@5 & +17.1 & $<10^{-3}$ & \textsc{CORE-T} & +3.7 & $2\!\times\!10^{-3}$ & \textsc{CORE-T} & +6.0 & $<10^{-3}$ & \textsc{CORE-T} \\
    DR@10 & +5.8 & $<10^{-3}$ & \textsc{CORE-T} & +0.9 & $.419$ & Tie & +9.2 & $<10^{-3}$ & \textsc{CORE-T} \\
    ReAct & +13.5 & $<10^{-3}$ & \textsc{CORE-T} & +1.4 & $.279$ & Tie & +2.4 & $.061$ & Tie \\
    REAR & +10.9 & $<10^{-3}$ & \textsc{CORE-T} & +7.1 & $<10^{-3}$ & \textsc{CORE-T} & +9.1 & $<10^{-3}$ & \textsc{CORE-T} \\
    \addlinespace[2pt]
    \midrule
    \multicolumn{10}{l}{\textbf{\textsc{Beaver}}} \\
    \midrule
    DR@5 & +2.4 & $.063$ & Tie & $-1.9$ & $.289$ & Tie & $-1.0$ & $.727$ & Tie \\
    DR@10 & +2.4 & $.063$ & Tie & +0.5 & $1.000$ & Tie & +1.0 & $.625$ & Tie \\
    ReAct & +2.4 & $.063$ & Tie & $-0.5$ & $1.000$ & Tie & $-1.0$ & $.727$ & Tie \\
    REAR & +1.4 & $.453$ & Tie & +1.9 & $.388$ & Tie & $-1.9$ & $.388$ & Tie \\
    \bottomrule
    \end{tabular}
    \end{adjustbox}
    \caption{\textbf{Exact McNemar tests for all 57 available EX$_{all}$ comparisons.}
    $\Delta$ is the EX$_{all}$ difference in percentage points (\textsc{CORE-T}$-$baseline). The winner is determined at $p<.05$.}
    \label{tab:paired-ex-all}
\end{table*}

\paragraph{Qualitative adjustment cases.}
We define a \emph{bridge table} as a gold table with retrieval relevance to the query $\le 0.5$. A success recovers all bridge tables dropped by selection, a soft failure recovers only a subset, and a failure adds noise without recovering a missing gold table. In the success case, adjustment restores the missing gold bridge table \texttt{frpm}, raising F1 from 50.0\% to 80.0\%. In the soft failure, it recovers \texttt{drivers} but still misses \texttt{driverstandings} and introduces the noisy table \texttt{laptimes}; F1 nevertheless rises from 57.1\% to 66.7\%. In the failure case, selection already contains both gold tables, but adjustment adds the unnecessary table \texttt{legalities}, lowering F1 from 66.7\% to 57.1\%. These cases illustrate when compatibility evidence fully repairs over-pruning, only partially repairs it, or introduces noise.

\begin{table*}[!t]
    \centering
    \small
    \setlength{\tabcolsep}{3pt}
    \renewcommand{\arraystretch}{1.12}
    \begin{adjustbox}{max width=\textwidth}
    \begin{tabular}{@{}l p{4.4cm} p{2.5cm} p{3.5cm} p{4.0cm} c@{}}
    \toprule
    \textbf{Case} & \textbf{Original query} & \textbf{Gold tables} & \textbf{Before adjustment} & \textbf{After adjustment} & \textbf{F1 (\%)} \\
    \midrule
    Success & Name schools in Riverside which the average of average math score for SAT is grater than 400, what is the funding type of these schools? & \makecell[l]{\texttt{frpm},\\\texttt{satscores}} & \makecell[l]{\texttt{satscores},\\\texttt{schools}} & \makecell[l]{\texttt{satscores}, \texttt{schools},\\\texttt{frpm}} & $50.0\!\rightarrow\!80.0$ \\
    Soft failure & How many times did Michael Schumacher won from races hosted in Sepang International Circuit? & \makecell[l]{\texttt{circuits}, \texttt{drivers},\\\texttt{driverstandings},\\\texttt{races}} & \makecell[l]{\texttt{circuits}, \texttt{races},\\\texttt{results}} & \makecell[l]{\texttt{circuits}, \texttt{races},\\\texttt{results}, \texttt{drivers},\\\texttt{laptimes}} & $57.1\!\rightarrow\!66.7$ \\
    Failure & What's the French name of the set of cards with ``Tendo Ice Bridge'' is in? & \makecell[l]{\texttt{cards},\\\texttt{set\_translations}} & \makecell[l]{\texttt{cards}, \texttt{foreign\_data},\\\texttt{set\_translations}, \texttt{sets}} & \makecell[l]{\texttt{cards}, \texttt{foreign\_data},\\\texttt{set\_translations}, \texttt{sets},\\\texttt{legalities}} & $66.7\!\rightarrow\!57.1$ \\
    \bottomrule
    \end{tabular}
    \end{adjustbox}
    \caption{\textbf{Representative success and failure modes of table adjustment.}
    The success case restores the missing bridge table; the soft failure recovers only part of the missing join path while adding noise; and the failure adds noise when selection already contains all gold tables.}
    \label{tab:qualitative-adjustment-cases}
\end{table*}

\paragraph{Heuristic error categories.}
We also perform an automatic, heuristic error analysis on \textsc{Bird} (n=1534) for one example configuration (\textsc{CORE-T} with Llama-3.1-8B-Instruct as table selector and Llama-3.2-3B-Instruct as SQL generator) and compare against \textsc{ARM} using the same configuration. We report two \emph{retrieval/selection} categories and two \emph{SQL-generation} categories. For each category, we report \emph{counts} and the corresponding \emph{rate over all queries} (count$/1534$) in Table~\ref{tab:error-analysis-bird}.

\paragraph{Retrieval/selection categories.}
Let $T_{\text{pred}}$ be the final selected table set and $T_{\text{gold}}$ the gold tables.
We mark:
(i) \textbf{Recall issue} if $T_{\text{gold}} \nsubseteq T_{\text{pred}}$ (at least one required table is missing);
(ii) \textbf{Precision issue} if $T_{\text{pred}} \setminus T_{\text{gold}} \neq \emptyset$ (any extra table is included).
These categories are not mutually exclusive.

\paragraph{SQL-generation categories.}
To separate SQL-generation errors from retrieval errors, we analyze only cases where the
generated SQL references \emph{all and only} the gold tables (so missing-table effects do not apply). For queries that still fail execution accuracy under this setting, we label:
(i) \textbf{Schema linking} (wrong columns used despite correct tables),
(ii) \textbf{Formatting} (value-format mismatch, e.g., dates).
Labels are applied heuristically and can overlap; schema-linking and formatting are each detected separately using SQLite error-string patterns.


\section{Additional Ablation Results}
\label{app:ablation-extra}

\begin{table*}[!t]
    \centering
    \footnotesize
    \setlength{\tabcolsep}{3pt}
    \renewcommand{\arraystretch}{1.05}
    \begin{adjustbox}{max width=\textwidth}
    \begin{tabular}{l l l c c c c c c}
    \toprule
    \textbf{Selector LLM} & \textbf{SQL Generator} & \textbf{Dataset} &
    \multicolumn{3}{c}{\textbf{EX$_{MT}$}} &
    \multicolumn{3}{c}{\textbf{EX$_{\text{all}}$}} \\
    \cmidrule(lr){4-6}\cmidrule(l){7-9}
    & & &
    \textbf{DR@10} & \textbf{+Sel.} & \textbf{+Adj.} &
    \textbf{DR@10} & \textbf{+Sel.} & \textbf{+Adj.} \\
    \midrule

    \multirow{12}{*}{\textbf{Qwen-2.5-7B}} &
    \multirow{4}{*}{\textbf{Llama-3.2-3B}} &
    \textsc{Bird} & 12.7 & \textbf{17.7} & \underline{17.2} & 12.8 & \textbf{19.3} & \underline{18.5} \\
    & & \textsc{Spider} & 35.3 & \underline{36.8} & \textbf{37.5} & 35.5 & \textbf{44.6} & \underline{41.0} \\
    & & \textsc{MMQA} & 13.4 & \textbf{21.6} & \underline{19.3} & 13.4 & \textbf{21.6} & \underline{19.2} \\
    & & \textsc{Beaver} & \underline{0.0} & \textbf{2.4} & \textbf{2.4} & \underline{0.0} & \textbf{2.4} & \textbf{2.4} \\
    \cmidrule(lr){2-9}
    & \multirow{4}{*}{\textbf{Gemma-3-4B}} &
    \textsc{Bird} & 12.9 & \underline{18.9} & \textbf{19.5} & 18.4 & \underline{24.1} & \textbf{24.6} \\
    & & \textsc{Spider} & \underline{36.2} & \textbf{44.4} & \textbf{44.4} & 51.2 & \textbf{54.1} & \underline{53.6} \\
    & & \textsc{MMQA} & 16.0 & \textbf{26.0} & \underline{25.2} & 16.0 & \textbf{26.0} & \underline{25.2} \\
    & & \textsc{Beaver} & \underline{1.5} & \textbf{2.4} & \textbf{2.4} & \underline{1.9} & \textbf{2.9} & \textbf{2.9} \\
    \cmidrule(lr){2-9}
    & \multirow{4}{*}{\textbf{GPT-4o-mini}} &
    \textsc{Bird} & \underline{37.9} & 35.9 & \textbf{38.6} & \underline{42.6} & 40.9 & \textbf{43.2} \\
    & & \textsc{Spider} & \textbf{57.7} & 54.0 & \underline{55.3} & \textbf{66.9} & 65.4 & \underline{65.8} \\
    & & \textsc{MMQA} & \underline{35.1} & 34.8 & \textbf{36.1} & \underline{35.1} & 34.8 & \textbf{36.0} \\
    & & \textsc{Beaver} & \underline{5.3} & \textbf{5.8} & \textbf{5.8} & \underline{5.7} & \textbf{6.2} & \textbf{6.2} \\

    \midrule

    \multirow{12}{*}{\textbf{Llama-3.1-8B}} &
    \multirow{4}{*}{\textbf{Llama-3.2-3B}} &
    \textsc{Bird} & 2.3 & \underline{15.3} & \textbf{15.7} & 3.3 & \underline{16.4} & \textbf{16.6} \\
    & & \textsc{Spider} & 8.1 & \underline{34.2} & \textbf{34.4} & 12.8 & \textbf{35.5} & \underline{34.5} \\
    & & \textsc{MMQA} & 1.3 & \textbf{20.3} & \underline{18.9} & 1.3 & \textbf{20.3} & \underline{18.8} \\
    & & \textsc{Beaver} & \underline{0.5} & \textbf{1.0} & \underline{0.5} & \underline{0.5} & \textbf{1.4} & \underline{0.5} \\
    \cmidrule(lr){2-9}
    & \multirow{4}{*}{\textbf{Gemma-3-4B}} &
    \textsc{Bird} & 11.3 & \underline{15.5} & \textbf{16.6} & 17.2 & \underline{21.6} & \textbf{21.9} \\
    & & \textsc{Spider} & 33.6 & \underline{41.2} & \textbf{43.8} & 49.7 & \underline{53.1} & \textbf{54.2} \\
    & & \textsc{MMQA} & 19.3 & \textbf{25.8} & \underline{22.8} & 19.5 & \textbf{25.8} & \underline{22.8} \\
    & & \textsc{Beaver} & 3.4 & \textbf{4.4} & \underline{3.9} & 3.3 & \textbf{4.3} & \underline{3.8} \\
    \cmidrule(lr){2-9}
    & \multirow{4}{*}{\textbf{GPT-4o-mini}} &
    \textsc{Bird} & 37.5 & \underline{37.8} & \textbf{38.6} & \underline{42.3} & \underline{42.3} & \textbf{43.0} \\
    & & \textsc{Spider} & \textbf{57.7} & 55.3 & \underline{56.9} & \underline{66.2} & \underline{66.2} & \textbf{66.7} \\
    & & \textsc{MMQA} & 34.7 & \textbf{36.9} & \underline{35.8} & 34.8 & \textbf{36.9} & \underline{35.8} \\
    & & \textsc{Beaver} & 3.9 & \underline{4.9} & \textbf{5.3} & 4.3 & \underline{5.3} & \textbf{5.7} \\

    \bottomrule
    \end{tabular}
    \end{adjustbox}

    \caption{\textbf{Step-wise EX trajectory across DR@10 $\rightarrow$ +Selection $\rightarrow$ +Adjustment.}
    Downstream execution accuracy for multi-table queries (EX$_{MT}$, $\ge 2$ gold tables) and all queries (EX$_{all}$) across selector LLMs and SQL generators on \textsc{Bird}, \textsc{Spider}, \textsc{MMQA}, and \textsc{Beaver}. Bold indicates the highest stage result and underline the second best within each SQL-generator/selector/dataset/metric block.}
    \label{tab:ablation-ex-trajectory-combined}
\end{table*}

\paragraph{Interpreting the retrieval trade-off.}\label{par:ablation-retrieval-tradeoff}
Selection and Adjustment serve complementary objectives. Selection removes distractors and is therefore the primary driver of compactness and F1, whereas Adjustment preserves the selected tables and restores strongly compatible candidates to repair over-pruning. For example, Adjustment increases PR from 79.0 to 87.0 on \textsc{Bird} with Qwen and from 94.4 to 96.6 on \textsc{Spider} with Llama, while adding only 0.7 tables on average in each case. Missing and extra tables also have different downstream costs: a single missing required table can make correct multi-table SQL impossible, whereas an extra table may be tolerated but can introduce ambiguity. Consequently, neither F1 nor PR alone determines whether Adjustment is beneficial; the downstream benefit of its PR gain must outweigh the cost of additional distractors.

\paragraph{Stage-wise execution trajectories.}\label{par:ablation-stagewise-ex}
Table~\ref{tab:ablation-ex-trajectory-combined} reports stage-wise EX trajectories (DR@10 $\rightarrow$ +Selection $\rightarrow$ +Adjustment) for all SQL generators and selector LLMs. At each stage, the SQL generator is conditioned on the tables returned by that stage.
For EX$_{MT}$, the largest gains over DR@10 occur with smaller SQL generators. On \textsc{Spider}, using Llama-3.1-8B as the selector, the full pipeline raises EX$_{MT}$ for Llama-3.2-3B from 8.1 to 34.4, a 26.3-point gain, with Selection accounting for most of the increase (8.1$\rightarrow$34.2). On \textsc{MMQA}, EX$_{MT}$ for Gemma-3-4B with Qwen selection similarly increases from 16.0 to 25.2. The effect of Adjustment is more setting-dependent: it can further improve EX by restoring join paths (e.g., Gemma-3-4B on \textsc{Spider} with Llama selection: 41.2$\rightarrow$43.8; GPT-4o-mini on \textsc{Bird} with Qwen selection: 35.9$\rightarrow$38.6), but it can reduce EX when added tables introduce noise (e.g., Gemma-3-4B on \textsc{MMQA} with Llama selection: 25.8$\rightarrow$22.8). EX$_{\text{all}}$ follows a similar recall--precision trade-off: Selection usually accounts for most of the gain, whereas Adjustment helps when it restores missing join context but can hurt when the expanded schema contains false positives that make SQL generation more difficult.


\begin{figure*}[!tbp]
    \centering
    \begin{tikzpicture}
    
    \begin{groupplot}[
      group style={group size=3 by 1, horizontal sep=1.5cm},
      xbar,
      height=5.2cm,
      width=0.33\linewidth,
      xmin=0, xmax=65,
      xtick={0,10,20,30,40,50,60},
      symbolic y coords={Bird,Spider,MMQA,Beaver},
      ytick=data,
      y dir=reverse,
      enlarge y limits=0.25,
      axis x line*=bottom,
      axis y line*=left,
      axis line style={black},
      tick align=outside,
      xmajorgrids=true,
      major grid style={draw=black!15, line width=0.3pt},
      ymajorgrids=false,
      bar shift auto,
      tick label style={font=\scriptsize},
      yticklabel style={font=\scriptsize},
      xlabel style={font=\scriptsize},
      title style={font=\scriptsize, at={(0.5,0.95)}, anchor=south},
      point meta=x,
      nodes near coords,
      every node near coord/.append style={
        font=\scriptsize,
        text=black,
        anchor=west,
        xshift=0.5pt,
        yshift=1.2pt,
        /pgf/number format/fixed,
        /pgf/number format/precision=1,
      },
      clip=false,
    ]
    
    \nextgroupplot[
      title={Llama-3.2-3B-Instruct},
    ]
    \addplot+[xbar, bar width=8.2pt, fill=tabfactblue!100, draw=none, area legend, bar shift=+3.0pt]
      coordinates {(2.3,Bird) (8.1,Spider) (1.3,MMQA) (0.5,Beaver)};
    \addplot+[xbar, bar width=8.2pt, fill=wikitqgold!100, draw=none, area legend, bar shift=-3.0pt]
      coordinates {(15.7,Bird) (34.4,Spider) (18.9,MMQA) (0.5,Beaver)};
    
    \nextgroupplot[
      title={Gemma-3-4B-Instruct},
      xlabel={Execution Accuracy (EX$_{\ge 2T}$)},
    ]
    \addplot+[xbar, bar width=8.2pt, fill=tabfactblue!100, draw=none, area legend, bar shift=+3.0pt]
      coordinates {(11.3,Bird) (33.6,Spider) (19.3,MMQA) (3.4,Beaver)};
    \addplot+[xbar, bar width=8.2pt, fill=wikitqgold!100, draw=none, area legend, bar shift=-3.0pt]
      coordinates {(16.6,Bird) (43.8,Spider) (22.8,MMQA) (3.9,Beaver)};
    
    \nextgroupplot[
      title={GPT-4o-mini},
    ]
    \addplot+[xbar, bar width=8.2pt, fill=tabfactblue!100, draw=none, area legend, bar shift=+3.0pt]
      coordinates {(37.5,Bird) (57.7,Spider) (34.7,MMQA) (3.9,Beaver)};
    \addplot+[xbar, bar width=8.2pt, fill=wikitqgold!100, draw=none, area legend, bar shift=-3.0pt]
      coordinates {(38.6,Bird) (56.9,Spider) (35.8,MMQA) (5.3,Beaver)};
    
    \end{groupplot}
    
    \node[anchor=north] at ($(group c1r1.south)!0.5!(group c3r1.south) + (0,-0.95cm)$) {%
      \scriptsize
      \begin{tikzpicture}[baseline]
        \matrix[column sep=5pt, row sep=0pt, nodes={inner sep=0pt, outer sep=0pt}]{
          \node{\tikz\fill[tabfactblue!100] (0,0) rectangle (0.55,0.15);}; &
          \node{DR@10}; &
          \node{\tikz\fill[wikitqgold!100] (0,0) rectangle (0.55,0.15);}; &
          \node{\textsc{CORE-T}}; \\
        };
      \end{tikzpicture}%
    };
    
    \end{tikzpicture}
    \caption{\textbf{Ablation (EX$_{MT}$).}
    We compare using only the first stage of our pipeline (DR@10) vs.\ full \textsc{CORE-T} pipeline.
    All settings use UAE-Large-V1 embeddings and Llama-3.1-8B-Instruct as the table selector, varying only the SQL generator.}
    \label{fig:em-ge2t-ablation}
\end{figure*}

\begin{figure*}[!tbp]
    \centering
    \begin{tikzpicture}
    
    \begin{groupplot}[
      group style={group size=3 by 1, horizontal sep=1.5cm},
      xbar,
      height=5.2cm,
      width=0.33\linewidth,
      xmin=0, xmax=70,
      xtick={0,10,20,30,40,50,60,70},
      symbolic y coords={Bird,Spider,MMQA,Beaver},
      ytick=data,
      y dir=reverse,
      enlarge y limits=0.25,
      axis x line*=bottom,
      axis y line*=left,
      axis line style={black},
      tick align=outside,
      xmajorgrids=true,
      major grid style={draw=black!15, line width=0.3pt},
      ymajorgrids=false,
      bar shift auto,                  
      tick label style={font=\scriptsize},
      yticklabel style={font=\scriptsize},
      xlabel style={font=\scriptsize},
      title style={font=\scriptsize, at={(0.5,0.95)}, anchor=south},
      point meta=x,
      nodes near coords,
      every node near coord/.append style={
        font=\scriptsize,
        text=black,
        anchor=west,
        xshift=0.5pt,
        /pgf/number format/fixed,
        /pgf/number format/precision=1,
      },
      clip=false,
    ]
    
    \nextgroupplot[
      title={Llama-3.2-3B-Instruct},
    ]
    \addplot+[xbar, bar width=8.8pt, fill=tabfactblue!100, draw=none, area legend, bar shift=+4.5pt]
      coordinates {(3.3,Bird) (12.8,Spider) (1.3,MMQA) (0.5,Beaver)};
    \addplot+[xbar, bar width=8.8pt, fill=wikitqgold!100, draw=none, area legend, bar shift=-4.5pt]
      coordinates {(16.6,Bird) (34.5,Spider) (18.8,MMQA) (0.5,Beaver)};
    
    \nextgroupplot[
      title={Gemma-3-4B-Instruct},
      xlabel={Execution Accuracy (EX$_{all}$)},
    ]
    \addplot+[xbar, bar width=8.8pt, fill=tabfactblue!100, draw=none, area legend, bar shift=+4.5pt]
      coordinates {(17.2,Bird) (49.7,Spider) (19.5,MMQA) (3.3,Beaver)};
    \addplot+[xbar, bar width=8.8pt, fill=wikitqgold!100, draw=none, area legend, bar shift=-4.5pt]
      coordinates {(21.9,Bird) (54.2,Spider) (22.8,MMQA) (3.8,Beaver)};
    
    \nextgroupplot[
      title={GPT-4o-mini},
    ]
    \addplot+[xbar, bar width=8.8pt, fill=tabfactblue!100, draw=none, area legend, bar shift=+4.5pt]
      coordinates {(42.3,Bird) (66.2,Spider) (34.8,MMQA) (4.3,Beaver)};
    \addplot+[xbar, bar width=8.8pt, fill=wikitqgold!100, draw=none, area legend, bar shift=-4.5pt]
      coordinates {(43.0,Bird) (66.7,Spider) (35.8,MMQA) (5.7,Beaver)};
    
    \end{groupplot}
    
    \node[anchor=north] at ($(group c1r1.south)!0.5!(group c3r1.south) + (0,-0.95cm)$) {%
      \scriptsize
      \begin{tikzpicture}[baseline]
        \matrix[column sep=5pt, row sep=0pt, nodes={inner sep=0pt, outer sep=0pt}]{
          \node{\tikz\fill[tabfactblue!100] (0,0) rectangle (0.55,0.15);}; &
          \node{DR@10}; &
          \node{\tikz\fill[wikitqgold!100] (0,0) rectangle (0.55,0.15);}; &
          \node{\textsc{CORE-T}}; \\
        };
      \end{tikzpicture}%
    };
    
    \end{tikzpicture}
    
    \caption{\textbf{Ablation (EX$_{all}$).}
    We compare using only the first stage of our pipeline (DR@10) vs. the full
    \textsc{CORE-T} pipeline.
    All settings use UAE-Large-V1 for retrieval and Llama-3.1-8B-Instruct as the table
    selector, varying only the SQL generator.}
    \label{fig:em-all-ablation}
\end{figure*}

\clearpage
\onecolumn          
\raggedbottom       

\section{Prompts Used}
\label{app:prompts}

\begin{lstlisting}[style=prompt, caption={Prompt for table purpose generation}, label={lst:prompt-purpose}]
Given the following table, describe the purpose of this table in layman's terms in one paragraph. If you do not think the text is semantically meaningful, output None.
{table}
\end{lstlisting}

\begin{lstlisting}[style=prompt, caption={Prompt for table selection}, label={lst:prompt-selection}]
You are a SQL schema analyst.
Your task: From a set of retrieved tables, identify a comprehensive set of tables that are BOTH:
(1) Relevant to the given query, and  
(2) Compatible (joinable) with each other to answer the query.

IMPORTANT:  
- Do NOT aggressively eliminate tables.  
- If there is a reasonable probability that a table is relevant and compatible, keep it.  
- When uncertain, prefer to keep the table rather than remove it -- it is better to have slightly more tables than to risk removing a necessary one.  
- Only remove a table if it is clearly irrelevant or incompatible.

---

### Information Provided:
- **Query**: {query}
- **Tables**: {tables_content}  
  Each table includes:
  - Table name
  - Table header and sample content in markdown format (5 rows)
- **Compatibility analysis (restricted to valid key-foreign key pairs)**: {compatibility_analysis}  
  For each pair of tables, compatibility scores are included **only if** at least one column of the first table is completely unique and at least one column of the second table is a subset of it.  
  If no such relationship exists, that pair is omitted (since all scores would be zero).  

  For included pairs, the following metrics are provided:
    - `overall_compatibility`: Highest weighted score between all possible column pairs that satisfy the constraint: one column is unique, the other is a subset of it.
    - `best_join_columns`: The specific column pair with the highest overall compatibility score.

---

### Step-by-step reasoning policy (YOU MUST FOLLOW THIS ORDER):

**Step 1 - Understand the query**  
- Identify the core entities and relationships.  
- Determine what type of data is required to answer it.

**Step 2 - Evaluate individual table relevance**  
- Use table name, column names, and sample data to decide if each table is relevant.  
- When unsure, treat the table as potentially relevant.

**Step 3 - Evaluate pairwise compatibility**  
For each pair of retrieved tables:  
- Interpret the compatibility scores.  
- Cross-check with table semantics from names, sample values.  
- When in doubt about compatibility, keep the pair as potentially relevant.

**Step 4 - Group formation**  
- Form one or more groups of tables where all members are mutually joinable.  
- Groups must form connected join graphs (no isolated tables).  
- Prefer forming larger groups when there is uncertainty rather than splitting unnecessarily.

**Step 5 - Group selection**  
- Select the single most relevant and compatible group for the query.  
- High recall is as important as precision in this step -- include tables that are possibly relevant to ensure coverage.

---

### Output Format:
Return the output as valid JSON in the following format:

{{
  "overall_reasoning": "Your general approach and observations about the tables and query",
  "group_formation": {{
    "reasoning": "How groups were formed based on provided quantitative and qualitative information",
    "groups_formed": [
      {{
        "group_index": 0,
        "table_indices": [0, 1, 2],
        "group_description": "Description of what this group represents"
      }}
    ]
  }},
  "group_selection": {{
    "selected_group_index": 0,
    "reasoning": "Detailed explanation of why this group was selected for the query",
    "group_analysis": [
      {{
        "group_index": 0,
        "reasoning": "Why this group is/isn't suitable for the query"
      }}
    ]
  }}
}}

---

### Few-shot Example

**Example Input**:
Query:
"In campaigns with exactly 2 events, how many of the events have clicks equal to 0?"

Tables:
Table 0:
Table name: campaigns  
Example table content:
| campaign_id | owner_id | name              | created_at           | event_count |
|------------:|---------:|-------------------|----------------------|------------:|
| 10          | 1        | Winter Launch     | 2024-01-05 10:00:00  | 2           |
| 11          | 2        | Spring Promo      | 2024-02-10 09:30:00  | 1           |
| 12          | 1        | Summer Teaser     | 2024-03-01 12:15:00  | 2           |

Table 1:
Table name: campaign_events  
Example table content:
| event_id | campaign_id | event_type | clicks | impressions | created_at           |
|---------:|------------:|-----------|-------:|------------:|----------------------|
| 100      | 10          | email     | 0      | 500         | 2024-01-05 10:05:00  |
| 101      | 10          | banner    | 12     | 1000        | 2024-01-05 10:06:00  |
| 102      | 11          | email     | 5      | 300         | 2024-02-10 09:35:00  |
| 103      | 12          | social    | 0      | 800         | 2024-03-01 12:20:00  |
| 104      | 12          | banner    | 7      | 900         | 2024-03-01 12:21:00  |

Table 2:
Table name: cities  
Example table content:
| city_id | name    | country | population |
|--------:|---------|---------|-----------:|
| 1       | Berlin  | DE      | 3600000    |
| 2       | Munich  | DE      | 1500000    |
| 3       | Hamburg | DE      | 1800000    |

Compatibility analysis:
Pair (Table 0 <-> Table 1):
  overall_compatibility: 0.96
  best_join_columns: "campaign_id <-> campaign_id"

**Example Output**:
{{
  "overall_reasoning": "The query is about campaigns and their events. The 'campaigns' table holds campaign-level data including event_count, while 'campaign_events' holds per-event data including clicks and campaign_id for linking. The 'cities' table is unrelated to the query and has no compatible join key with the other tables.",
  "group_formation": {{
    "reasoning": "Formed one group with 'campaigns' and 'campaign_events' because they are both relevant to the query and strongly joinable via campaign_id <-> campaign_id. 'cities' is excluded due to lack of relevance and join compatibility.",
    "groups_formed": [
      {{
        "group_index": 0,
        "table_indices": [0, 1],
        "group_description": "Campaigns and their associated events, enabling filtering by event_count and counting events with clicks = 0."
      }}
    ]
  }},
  "group_selection": {{
    "selected_group_index": 0,
    "reasoning": "This group contains all and only the tables needed to answer the query: campaigns to identify those with exactly 2 events, and campaign_events to count events with clicks equal to 0.",
    "group_analysis": [
      {{
        "group_index": 0,
        "reasoning": "Fully suitable and sufficient for the query; no other table contributes necessary information."
      }}
    ]
  }}
}}
\end{lstlisting}

\begin{lstlisting}[style=prompt, caption={ReAct-style prompt for table retrieval}, label={lst:prompt-react}]
You are a table-retrieval ReAct agent. Your ONLY goal is to pick a
VERY HIGH-RECALL set of SQL tables required (or plausibly helpful) to answer the user's question.
Do NOT compute the answer.

You can call one tool: `table_search`. It returns 5 NEW candidate tables as JSON rows:
- table_index  (unique integer id and is stable per dataset build)
- table_name
- purpose
- table_markdown_content  (markdown with table name, column headers, and 5 sample rows)
(You must infer relevance and joinability from names, purposes, and columns shown.)

Recall-first rules (critical):
- Prefer **recall over precision**. If a table is plausibly useful (lookup, join bridge, calendar/date, entity master, hierarchy, mapping),
  **include it**, even if not strictly necessary.
- When the question names an entity (concerts, teams, categories), **include the entity master** and plausible **bridge/mapping** tables.
- If multiple tables could host a needed field or join (synonyms/overlaps like `songs` vs `tracks`, `date_dim` vs `calendar`), **keep both**.
- Keep **helper context** tables (calendar/date, region/geo, category/lookup, hierarchy) that could influence grouping/filters.
- Do **not** aim for a minimal set. Slight redundancy is acceptable. **When in doubt, include.**
- Target a recall-oriented set. **Deduplicate indices**; prefer cohesive families (master + bridge + lookups).

Iteration protocol:
1) Thought: infer entities, fields, joins, and helper lookups (no SQL, no answer).
2) Action: table_search
3) Observation: inspect candidates (table_index, table_name, purpose, table_markdown_content)

After every Observation, write:
Thought: summarize which fields/joins/helpers are now covered and if gaps/ambiguities remain.
Seen tables: [table_indices so far]
New tables this step: [table_indices discovered this step]
Coverage: fields=<yes/no> joins=<yes/no> helpers=<yes/no> gaps=<yes/no>

Hard stop rule:
- If this step discovered **NO new** table_index (Observation is []), STOP and output final JSON.
- Never repeat an identical Action Input. If you lack new terms, STOP.

Output when stopping (INDICES ONLY; deduped; recall-first):
Final Answer: {"relevant_tables":[0, 12, 44]}

Examples:

Example 1:
Question: Which authors spoke at any festival and what are their ages?
Thought: Need a mapping between authors and festivals, author demographics (age), and festival context. Venue details might be useful if location filters appear.
Action: table_search
Action Input: "author festival participation lineup mapping"
Observation: [
  {"table_index": 27, "table_name":"author_festival_link","purpose":"Author presence per festival (festival_id, author_id)","table_markdown_content":"... | festival_id | author_id | ..."},
  {"table_index": 22, "table_name":"festival","purpose":"Literary festivals with name, venue_id, year","table_markdown_content":"... | festival_id | name | venue_id | year | ..."},
  {"table_index": 18, "table_name":"venue","purpose":"Venue lookup","table_markdown_content":"... | venue_id | name | city | capacity | ..."},
  {"table_index": 44, "table_name":"ticket_sales","purpose":"Ticket sales per festival","table_markdown_content":"... | ticket_id | festival_id | price | ..."},
  {"table_index": 39, "table_name":"staff_profile","purpose":"Festival staff master","table_markdown_content":"... | staff_id | name | age | role | ..."}
]
Thought: Keep 27 (bridge), 22 (festival master), 18 (venue context may become relevant). Ticket sales and staff profiles are not needed for author ages.
Seen tables: [27, 22, 18]
New tables this step: [27, 22, 18]
Coverage: fields=partial (no author age yet) joins=yes helpers=yes gaps=yes
Action: table_search
Action Input: "author master id age biography country"
Observation: [
  {"table_index": 3, "table_name":"author","purpose":"Author master with id, name, age, country, etc.","table_markdown_content":"... | author_id | name | age | country | ..."},
  {"table_index": 8, "table_name":"book","purpose":"Books with title, release_year, author reference","table_markdown_content":"... | book_id | title | release_year | author_id | ..."},
  {"table_index": 29, "table_name":"series","purpose":"Series info","table_markdown_content":"... | series_id | title | start_year | ..."},
  {"table_index": 30, "table_name":"literary_prize","purpose":"Prizes awarded to authors or books","table_markdown_content":"... | prize_id | name | year | ..."},
  {"table_index": 25, "table_name":"reading_list","purpose":"Curated reading lists","table_markdown_content":"... | list_id | name | ..."}
]
Thought: Keep 3 (author ages). Also keep 8 (book) as plausible context if book metadata is referenced later. Others are less central now.
Seen tables: [27, 22, 18, 3, 8]
New tables this step: [3, 8]
Coverage: fields=yes joins=yes helpers=yes gaps=no
Final Answer: {"relevant_tables":[27, 22, 18, 3, 8]}

Example 2:
Question: Who is the youngest author and what is the publication year of their first book?
Thought: Need an author table (age or birth_year) and a book table (publication_year, author_id). Extra context like series or imprints may help interpret "first" publication.
Action: table_search
Action Input: "author age birth_year id name"
Observation: [
  {"table_index": 3,  "table_name":"author","purpose":"Author master incl. birth_year/age","table_markdown_content":"... | author_id | name | birth_year | age | ..."},
  {"table_index": 27, "table_name":"author_festival_link","purpose":"Author presence per festival","table_markdown_content":"... | festival_id | author_id | ..."},
  {"table_index": 41, "table_name":"author_award","purpose":"Awards received by authors","table_markdown_content":"... | award_id | author_id | year | ..."},
  {"table_index": 19, "table_name":"writing_group","purpose":"Writing groups","table_markdown_content":"... | group_id | name | ..."},
  {"table_index": 55, "table_name":"literary_agent","purpose":"Agents representing authors","table_markdown_content":"... | agent_id | name | ..."}
]
Thought: Keep 3. Others do not directly help determine the debut publication year.
Seen tables: [3]
New tables this step: [3]
Coverage: fields=partial joins=no helpers=no gaps=yes
Action: table_search
Action Input: "book title publication_year author_id debut earliest"
Observation: [
  {"table_index": 8,  "table_name":"book","purpose":"Books with title, publication_year, author_id","table_markdown_content":"... | book_id | title | publication_year | author_id | ..."},
  {"table_index": 29, "table_name":"series","purpose":"Series info (title, start_year)","table_markdown_content":"... | series_id | title | start_year | ..."},
  {"table_index": 46, "table_name":"imprint","purpose":"Publishing imprints","table_markdown_content":"... | imprint_id | name | ..."},
  {"table_index": 25, "table_name":"reading_list","purpose":"Reading lists","table_markdown_content":"... | list_id | name | ..."},
  {"table_index": 62, "table_name":"writing_session","purpose":"Writing sessions","table_markdown_content":"... | session_id | studio | date | ..."}
]
Thought: Keep 8 (publication_year + author_id). Also keep 29 (series) and 46 (imprint) as plausible helpers when reasoning about earliest works, for better recall.
Seen tables: [3, 8, 29, 46]
New tables this step: [8, 29, 46]
Coverage: fields=yes joins=yes helpers=yes gaps=no
Final Answer: {"relevant_tables":[3, 8, 29, 46]}

Example 3:
Question: For each cuisine in 2022, report total bookings and average party size by loyalty tier.
Thought: Need bookings (date, total_spend, party_size, guest_id), booking line items (booking_id, menu_item_id, quantity), menu items (menu_item_id, cuisine_id), cuisines (cuisine_name), and guest profiles (loyalty tier). A date dimension and payment/area lookup tables may help with time and context filters.
Action: table_search
Action Input: "booking table guest cuisine date amount loyalty tier"
Observation: [
  {"table_index": 70, "table_name":"dining_bookings","purpose":"Restaurant bookings with date and totals","table_markdown_content":"... | booking_id | booking_date | guest_id | party_size | total_spend | ..."},
  {"table_index": 71, "table_name":"booking_items","purpose":"Line items per booking","table_markdown_content":"... | line_id | booking_id | menu_item_id | quantity | line_total | ..."},
  {"table_index": 72, "table_name":"menu_items","purpose":"Menu item master incl. cuisine_id","table_markdown_content":"... | menu_item_id | name | cuisine_id | ..."},
  {"table_index": 73, "table_name":"cuisines","purpose":"Cuisine lookup","table_markdown_content":"... | cuisine_id | cuisine_name | ..."},
  {"table_index": 74, "table_name":"guest_profiles","purpose":"Guest master incl. loyalty_tier","table_markdown_content":"... | guest_id | loyalty_tier | city_code | ..."}
]
Thought: Keep all five: they give bookings, items, cuisine classification, and guest tiers. For recall, also include date and payment/geography helpers.
Seen tables: [70, 71, 72, 73, 74]
New tables this step: [70, 71, 72, 73, 74]
Coverage: fields=yes joins=yes helpers=partial gaps=maybe
Action: table_search
Action Input: "date dimension year month day payment log area lookup geography"
Observation: [
  {"table_index": 90, "table_name":"date_dimension","purpose":"Date dimension with year/month/day","table_markdown_content":"... | date_key | year | month | day | ..."},
  {"table_index": 77, "table_name":"payment_log","purpose":"Payment records per booking","table_markdown_content":"... | payment_id | booking_id | amount | method | ..."},
  {"table_index": 78, "table_name":"area_lookup","purpose":"Geographic areas","table_markdown_content":"... | area_id | name | ..."},
  {"table_index": 79, "table_name":"restaurant_branch","purpose":"Restaurant branches","table_markdown_content":"... | branch_id | name | area_id | ..."},
  {"table_index": 76, "table_name":"delivery_schedule","purpose":"Delivery schedule info","table_markdown_content":"... | schedule_id | booking_id | delivery_date | ..."}
]
Thought: Keep 90 (date_dimension), 77 (payment_log), and 78 (area_lookup) as plausible helpers. Restaurant branches and delivery schedule are less central to cuisine-level booking stats.
Seen tables: [70, 71, 72, 73, 74, 90, 77, 78]
New tables this step: [90, 77, 78]
Coverage: fields=yes joins=yes helpers=yes gaps=no
Final Answer: {"relevant_tables":[70, 71, 72, 73, 74, 90, 77, 78]}

Begin!

Question: {input}
{agent_scratchpad}
\end{lstlisting}

\begin{lstlisting}[style=prompt, caption={Prompt for SQL generation}, label={lst:prompt-sql-generation}]
You are an expert SQL query generator. Given the following tables and a natural language question, generate a precise SQL query that answers the question. The target dialect is SQLite.

AVAILABLE TABLES:
{schema_text}

External knowledge: {evidence}

QUESTION: {query}

INSTRUCTIONS:
1. Analyze the question carefully to understand what information is being requested.
2. Identify which tables and columns are needed from the available tables. Use only the provided tables and columns; never invent schema or values. Treat sample values in the schema as illustrative, not exhaustive.
3. Generate a syntactically correct SQL query that answers the question.
4. Use explicit JOIN ... ON ... with appropriate JOINs when multiple tables are needed; avoid cartesian products.
5. Apply proper filtering, grouping, ordering, LIMIT, DISTINCT, HAVING, and aggregates as required.
6. Use the exact column names as shown in the schema; qualify ambiguous column names with table names or aliases.
7. Be careful with column names that contain spaces--use backticks or quotes as needed.
8. Return ONLY the SQL query without any explanation, comments, or markdown formatting.

SQL QUERY:
\end{lstlisting}

\end{document}